\pdfoutput=1

\documentclass[11pt]{article}

\usepackage[]{EMNLP2022}

\usepackage{times}
\usepackage{latexsym}

\usepackage[T1]{fontenc}

\usepackage[utf8]{inputenc}

\usepackage{microtype}

\usepackage{inconsolata}

\usepackage{graphicx}
\usepackage{amsmath}
\usepackage{algorithm}
\usepackage{algpseudocode}
\usepackage{booktabs}
\usepackage{xspace}
\usepackage{hyperref}
\usepackage{multirow}
\usepackage{subfigure}
\usepackage{float}

\def\md{$m^{4}Adapter$\xspace}

\newcounter{notecounter}

\newcommand{\enoteson}{\long\gdef\enote##1##2{{
\stepcounter{notecounter}
{\large\bf \hspace{1cm}\arabic{notecounter} $<<<$ ##1: ##2 $>>>$\hspace{1cm}}}}}
\enoteson

\title{\boldmath $m^{4} Adapter$: Multilingual Multi-Domain Adaptation for Machine Translation with a Meta-Adapter}

\author{Wen Lai$^{1}$, Alexandra Chronopoulou$^{1,2}$, Alexander Fraser$^{1,2}$\\\\
	$^1$Center for Information and Language Processing, LMU Munich, Germany \\
    $^2$Munich Center for Machine Learning, Germany \\
     {\tt \{lavine, achron, fraser\}@cis.lmu.de}}

\begin{document}
\maketitle
\begin{abstract}
Multilingual neural machine translation models (MNMT) yield state-of-the-art performance when evaluated on data from a domain and language pair seen at training time.
However, when a MNMT model is used to translate under domain shift or to a new language pair, performance drops dramatically.
We consider a very challenging scenario: adapting the MNMT model both to a new domain and to a new language pair at the same time.
In this paper, we propose \md (\underline{M}ultilingual \underline{M}ulti-Domain Adaptation for \underline{M}achine Translation with a \underline{M}eta-\underline{Adapter}), which combines domain and language knowledge using meta-learning with adapters.
We present results showing that our approach is a parameter-efficient solution which effectively adapts a model to both a new language pair and a new domain, while outperforming other adapter methods.
An ablation study also shows that our approach more effectively transfers domain knowledge across different languages and language information across different domains.\footnote{Our source code is available at \url{https://github.com/lavine-lmu/m4Adapter}}
\end{abstract}

\section{Introduction}
Multilingual neural machine translation (MNMT; \citealp{johnson-etal-2017-googles,aharoni-etal-2019-massively,fan2021beyond}), uses a single model to handle translation between multiple language pairs.
There are two reasons why MNMT is appealing: first, it has been proved to be effective on transferring knowledge from high-resource languages to low-resource languages, especially in zero-shot scenarios \cite{gu-etal-2019-improved,zhang-etal-2020-improving}; second, it significantly reduces training and inference cost, as it requires training only a single multilingual model, instead of a separate model for each language pair.

Adapting MNMT models to multiple domains is still a challenging task, particularly when domains are distant to the domain of the training corpus.
One approach to address this is \textit{fine-tuning} the model on out-of-domain data for NMT \cite{freitag2016fast,dakwale2017finetuning}.
Another approach is to use lightweight, learnable units inserted between transformer layers, which are called \textit{adapters}  \cite{bapna-firat-2019-simple} for each new domain.
Similarly, there is research work on adapting MNMT models to a new language pair using fine-tuning \cite{neubig-hu-2018-rapid} and adapters \cite{bapna-firat-2019-simple,philip-etal-2020-monolingual,cooper-stickland-etal-2021-recipes}.

Although effective, the above approaches have some limitations: 
i) Fine-tuning methods require updating the parameters of the whole model for each new domain, which is costly; 
ii) when fine-tuning on a new domain, catastrophic forgetting  \cite{mccloskey1989catastrophic} reduces the performance on all other domains, and proves to be a significant issue when data resources are limited. 
iii) adapter-based approaches require training domain adapters for each domain and language adapters for all languages, which also becomes parameter-inefficient when adapting to a new domain and a new language because the parameters scale linearly with the number of domains and languages. 

In recent work,  \citet{cooper-stickland-etal-2021-multilingual} compose language adapters and domain adapters in MNMT and explore to what extent domain knowledge can be transferred across languages.
They find that it is hard to decouple language knowledge from domain knowledge and that adapters often cause the `off-target' problem (i.e., translating into a wrong target language \cite{zhang-etal-2020-improving}) when new domains and new language pairs are combined together.
They address this problem by using additional in-domain monolingual data to generate synthetic data (i.e., back-translation; \citealp{sennrich-etal-2016-improving}) and randomly dropping some domain adapter layers (AdapterDrop; \citealp{ruckle-etal-2021-adapterdrop}).

Motivated by \citet{cooper-stickland-etal-2021-multilingual}, we consider a challenging scenario: adapting a MNMT model to multiple new domains and new language directions simultaneously in low-resource settings without using extra monolingual data for back-translation.
This scenario could arise when one tries to translate a domain-specific corpus with a commercial translation system.
Using our approach, we adapt a model to a new domain and a new language pair using just 500 domain- and language-specific sentences.

To this end, we propose \md (\underline{M}ultilingual \underline{M}ulti-Domain Adaptation for \underline{M}achine Translation with \underline{M}eta-\underline{Adapter}), which facilitates the transfer between different domains and languages using meta-learning \cite{finn2017model} with adapters.
Our hypothesis is that we can formulate the task, which is to adapt to new languages and domains, as a multi-task learning problem (and denote it as  \textit{$D_i$-$L_1$-$L_2$}, which stands for translating from a language $L_1$ to a language $L_2$ \textit{in a specific domain} $D_i$).
Our approach is two-step: initially, we perform meta-learning with adapters to efficiently learn parameters in a shared representation space across multiple tasks using a small amount of training data (5000 samples); we refer to this as the meta-training step. Then, we fine-tune the trained model to a new domain and language pair simultaneously using an even smaller dataset (500 samples); we refer to this as the meta-adaptation step.

In this work, we make the following contributions:
i) We present \md, a meta-learning approach with adapters that can easily adapt to new domains and languages using a single MNMT model. Experimental results show that \md outperforms strong baselines.
ii) Through an ablation study, we show that using \md, domain knowledge can be transferred across languages and language knowledge can also be transferred across domains without using target-language monolingual data for back-translation (unlike the work of \citealp{cooper-stickland-etal-2021-multilingual}).
iii) To the best of our knowledge, this paper is the first work to explore meta-learning for MNMT adaptation.

\section{Related Work}
\noindent \textbf{Domain Adaptation in NMT.}
Existing work on domain adaptation for machine translation can be categorized into two types: \textit{data-centric} and \textit{model-centric} approaches \cite{chu-wang-2018-survey}.
The former focus on maximizing the use of in-domain monolingual, synthetic, and parallel data \cite{domhan-hieber-2017-using,park2017building,van-der-wees-etal-2017-dynamic}, while the latter design specific training objectives, model architectures or decoding algorithms for domain adaptation \cite{khayrallah-etal-2017-neural,gu-etal-2019-improved,park-etal-2022-dalc}. 
In the case of MNMT, adapting to new domains is more challenging because it needs to take into account transfer between languages \cite{chu2019multilingual,cooper-stickland-etal-2021-multilingual}.

\noindent \textbf{Meta-Learning for NMT.}
Meta-learning \cite{finn2017model}, which aims to learn a generally useful model by training on a distribution of tasks, is highly effective for fast adaptation and has recently been shown to be beneficial for many NLP tasks \cite{lee2022meta}.
\citet{gu-etal-2018-meta} first introduce a model-agnostic meta-learning algorithm (MAML; \citealp{finn2017model}) for low-resource machine translation. \citet{sharaf-etal-2020-meta}, \citet{zhan2021meta} and \citet{lai-etal-2022-improving-domain}
formulate domain adaptation for NMT as a meta-learning task, and show effective performance on adapting to new domains.
Our approach leverages meta-learning to adapt a MNMT model to a \textit{new domain} and to a \textit{new language pair} at the same time. 

\noindent \textbf{Adapters for NMT.}
\citet{bapna-firat-2019-simple} train \textit{language-pair} adapters on top of a pre-trained generic MNMT model, in order to recover lost performance on high-resource language pairs compared to bilingual NMT models.
\citet{philip-etal-2020-monolingual} train adapters for \textit{each language} and show that adding them to a trained model improves the performance of zero-shot translation.
\citet{chronopoulou2022language} train adapters for \textit{each language family} and show promising results on multilingual machine translation.
\citet{cooper-stickland-etal-2021-recipes} train \textit{language-agnostic} adapters to efficiently fine-tune a pre-trained model for many language pairs. 
More recently, \citet{cooper-stickland-etal-2021-multilingual} stack language adapters and domain adapters on top of an MNMT model and they conclude that it is not possible to transfer domain knowledge across languages, except by employing back-translation which requires significant in-domain resources.
In this work, we introduce adapters into the meta-learning algorithm and show that this approach permits transfer between domains and languages.

Our work is mostly related to \citet{cooper-stickland-etal-2021-multilingual}, however we note several differences:
i) we study a more realistic scenario: the corpus of each domain and language pair is low-resource (i.e., the meta-training corpus in each domain for each language pair is limited to 5000 sentences and the fine-tuning corpus to 500 sentences), which is easier to obtain;
ii) our approach can simultaneously adapt to new domains and new language pairs \textit{without using back-translation}.
iii) we also show that \md can transfer domain information across different languages and language knowledge across different domains through a detailed ablation analysis.

\section{Method}
Our goal is to efficiently adapt an MNMT model to new domains and languages.
We propose a novel approach, \md, which formulates the multilingual multi-domain adaptation task as a multi-task learning problem.
To address it, we propose a 2-step approach, which combines \textit{meta-learning} and \textit{meta-adaptation} with adapters. Our approach permits sharing parameters across different tasks. The two steps are explained in Subsections \ref{method:meta_training} and \ref{method:meta_adapt}.

\subsection{Meta-Training}
\label{method:meta_training}
The goal of meta-learning is to obtain a model that can easily adapt to new tasks.
To this end, we meta-train adapters in order to find a good initialization of our model's parameters using a small training dataset of source tasks $\left\{\mathcal{T}_{1}, \ldots, \mathcal{T}_{t}\right\}$.

We first select $m$ tasks, as we describe in $\S$~\ref{method:tasks}. 
Then, for each of the $m$ sampled tasks, we sample $n$ examples. We explain the task sampling strategy in $\S$~\ref{method:task_sampling}. This way, we set up the \textit{m-way-n-shot} task.  After setting up the task, we use a meta-learning algorithm, which we describe in $\S$ \ref{method:learnig_stratigies}, to meta-learn the parameters of the adapter layers. The architecture of the adapters and their  optimization objective are presented in $\S$~\ref{method:meta-adapter}. 
Algorithm~\ref{alg:meta_training} details the meta-training process of our approach.

\subsubsection{Task Definition}
\label{method:tasks}
Motivated by the work of \citet{tarunesh-etal-2021-meta}, where a multilingual multi-task NLP task is regarded as a Task-Language pair (TLP), 
we address multilingual multi-domain translation as a multi-task learning problem. Specifically, a translation task in a specific textual domain corresponds to a Domain-Language-Pair (\textbf{DLP}). For example, an English-Serbian translation task in the `Ubuntu' domain is denoted as a DLP `Ubuntu-en-sr'. 
Given $d$ domains and $l$ languages, we have $d \cdot l \cdot (l-1)$ tasks of this form.\footnote{Given $l$ languages, we focus on complete translation between $l\cdot(l-1)$ language directions.}
We denote the proportion of the dataset size of all DLPs for the $i^{th}$ DLP as $s_{i} = |\mathcal{D}_{train}^{i}|/\big(\sum_{a = 1}^{n} |\mathcal{D}_{train}^{a}|\big)$, where $s_i$ will be used in temperature-based sampling (see more details in $\S$~\ref{method:task_sampling}).
The probability of sampling a batch from the $i^{th}$ DLP during meta-training is denoted as $P_{\mathcal{D}}(i)$. The distribution over all DLPs, is a multinomial (which we denote as $\mathcal{M}$) over $P_{\mathcal{D}}(i)$: $\mathcal{M}\sim P_{\mathcal{D}}(i)$.

\subsubsection{Task Sampling}
\label{method:task_sampling}
Given $d$ domains and $l$ languages, we sample some DLPs per batch among all $d \cdot l \cdot (l-1)$ tasks.
We consider a standard \textit{m-way-n-shot} meta-learning scenario: assuming access to $d \cdot l \cdot (l-1)$ DLPs, a \textit{m-way-n-shot} task is created by first sampling $m$ DLPs ($m \ll l \cdot (l-1)$); then, for each of the $m$ sampled DLPs, ($n + q$) examples of each DLP are selected; the $n$ examples for each DLP serve as the support set to update the parameter of pre-trained model, while $q$ examples constitute the query set to evaluate the model.

Task sampling is an essential step for meta-learning.
Traditional meta-learning methods sample the tasks uniformly \cite{sharaf-etal-2020-meta}, through ordered curriculum \cite{zhan2021meta}, or dynamically adjust the sampled dataset according to the model parameters (parameterized sampling strategy, \citealp{tarunesh-etal-2021-meta}).
We do not employ these strategies for the following reasons: i) sampling uniformly is simple but does not consider the distribution of the unbalanced data; ii) Although effective, curriculum-based and parameterized sampling consider features of all $d\cdot l\cdot(l-1)$ DLPs. Because of this, the amount of DLPs is growing exponentially with the number of languages and domains.
In contrast, we follow a temperature-based heuristic sampling strategy \cite{aharoni-etal-2019-massively}, which defines the probability of any dataset as a function of its size.
Specifically, given $s_{i}$ as the percentage of 
the
$i^{th}$ DLP in all DLPs, we compute the following probability of the $i^{th}$ DLP to be sampled:
$$
P_{\mathcal{D}}(i)=s_{i}^{1 / \tau} /\left(\sum_{a=1}^{n} s_{a}^{1 / \tau}\right)
$$
where $\tau$ is a temperature parameter. $\tau=1$ means that each DLP is sampled in proportion to the size of the corresponding dataset. $\tau\rightarrow\infty$ refers to sampling DLPs uniformly.

\begin{figure}[t]
\centering
	\includegraphics[width=1.0\linewidth]{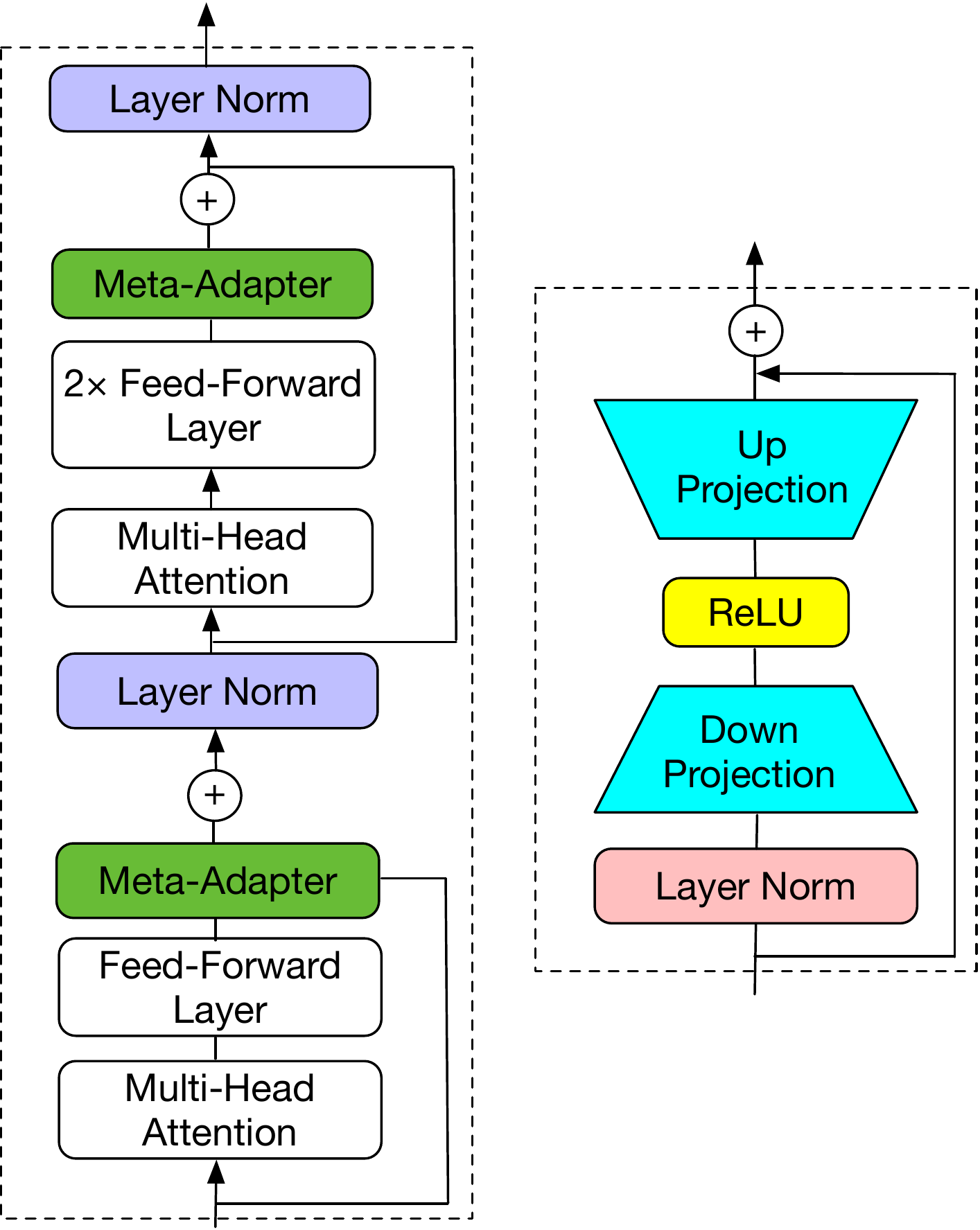}
	\caption{\label{fig:models} \md architecture.}
\end{figure}

\subsubsection{Meta-Learning Algorithm}
\label{method:learnig_stratigies}
Given $\theta$ as the parameters of the pre-trained model, $\psi$ as the parameters of the adapters, MAML aims to minimize the following objective:
$$
\min _{\psi} \sum_{\mathcal{T} _{i} \sim \mathcal{M}} \mathcal{L}_{i}\left(U_{i}^{k}(\theta,\psi)\right)
$$
where $\mathcal{M}$ is the multinomial distribution over DLPs, $\mathcal{L}_{i}$ is the loss function and $U_{i}^{k}$ is a function which keeps $\theta$ frozen and only returns $\psi$ after $k$ gradient updates calculated on batches sampled from $\mathcal{T}_{i}$.
Note that, to minimize this goal, the traditional MAML algorithm requires computing gradients of the form $\frac{\partial}{\partial \psi} U_{i}^{k}(\psi)$, which leads to the costly computation of second-order derivatives.
To this end, we follow \textit{Reptile} \cite{nichol2018first}, an alternative first-order meta-learning algorithm that uses a simple update rule:
$$
\psi \leftarrow \psi + \beta \frac{1}{|\{\mathcal{T}_{i}\}|} \sum_{\mathcal{T}_{i} \sim \mathcal{M}} (\psi^{(k)}_{i} - \psi)
$$
where $\psi_{i}^{(k)}$ is $U^{k}_{i} (\theta,\psi)$ and $\beta$ is a hyper-parameter.
Despite its simplicity, it was recently shown that Reptile is at least as effective as MAML in terms of performance \cite{dou-etal-2019-investigating}. We therefore employ Reptile for meta-learning in our experiments. 

\subsubsection{Meta-Adapter}
\label{method:meta-adapter}
Adapters
\cite{swietojanski2014learning, vilar-2018-learning, houlsby2019parameter} are lightweight feed-forward modules. They are described by the following Equation: $W_{\text {up }} f\left(W_{\text {down}} \operatorname{LN} \left(\mathbf{h}\right)\right) + \mathbf{h}$. An adapter consists of a layer normalization $\operatorname{LN}(\cdot)$ \cite{ba2016layer} of the input $\mathbf{h}$, which is passed to a down-projection $W_{\text {down}}\in R^{z \times d}$, a non-linear activation $f(\cdot)$ (in our case, ReLU) and an up-projection $W_{\text {up}}\in R^{d \times z}$, where $d$ is the bottleneck dimension of the adapter module and the only tunable hyperparameter. The output is combined with a residual connection. 
Adapters are added between sub-layers of a pre-trained Transformer \cite{attention} model (see the right part of Figure~\ref{fig:models}), usually after the feed-forward layer.

Using adapters is appealing for multiple reasons: i) we only update the adapter parameters $\psi$ during the whole fine-tuning process, which makes training faster especially for large pre-trained models; ii) they obtain a performance comparable to that of traditional fine-tuning.
However, as the adapter parameters $\psi$ are randomly initialized they may not perform well in the few-shot setting.
Moreover, adding a new set of adapters for each domain or language pair \cite{bapna-firat-2019-simple,cooper-stickland-etal-2021-multilingual} quickly becomes inefficient when we need to adapt to many new domains and language pairs.
To address this problem, we propose training a \textit{Meta-Adapter}, which inserts adapter layers into the meta-learning training process (see the left part of Figure~\ref{fig:models}). 
Different from the traditional adapter training process, we only need to train a single meta-adapter to adapt to all new language pairs and domains. 

Let $\theta$ denote the parameters of the pre-trained model and $\psi$ the parameters of the adapter. Given a target task $\mathcal{T}$ in the domain $\mathcal{D}_{\mathcal{T}}$ and a loss function $\mathcal{L}_{\mathcal{T}}(\cdot)$, we train a meta-adapter to minimize the following objective through gradient descent:
$$\min _{\psi} \mathcal{L}_{\mathcal{T}}\left(\theta, \psi ; \mathcal{D}_{\mathcal{T}}\right)$$ where the parameters of pre-trained model $\theta$ are frozen and the adapter parameters  $\psi$ are randomly initialized, leading to a size of $\psi \ll \theta$. This makes our approach more efficient than meta-learning an entire model (see more details in Section~\ref{analysis:efficiency}).

\subsection{Meta-Adaptation}
\label{method:meta_adapt}
After the meta-training phase, the parameters of the adapter are fine-tuned to adapt to new tasks (as both the domain and language pair of interest are not seen during the meta-training stage) using a small amount of data to simulate a low-resource scenario.

We find that this step is essential to our approach, as it permits adapting the parameters of the meta-learned model to the domain and language pair of interest. This step uses a very small amount of data (500 samples), which we believe could realistically be available for each DLP.

\begin{algorithm}[t]
\textbf{Input:} ${\mathcal{D}}_{train}$ set of DLPs for meta training; Pre-trained MNMT model $\theta$
\caption{$m^{4}Adapter$ (Multilingual Multi-Domain Adaptation with Meta-Adapter)}\label{alg:meta_training}
\begin{algorithmic}[1]
\State \textbf{Initialize} $P_{D}(i)$ based on temperature sampling
    \While{not converged}
    \State $\triangleright$ \textit{Perform Reptile Updates}
    \State Sample $m$ DLPs $\mathcal{T}_{1}, \mathcal{T}_{2}, \dots, \mathcal{T}_{m}$ from $\mathcal{M}$
    \For{i = 1,2,\dots,m}
        \State $\psi_i^{(k)} \leftarrow U_{i}^{k}(\theta,\psi),$ denoting $k$ gradient \\
        \quad\quad\quad updates from $\psi$ on batches of DLP $\mathcal{T}_{i}$ \\
        \quad\quad\quad while keeping $\theta$ frozen \label{algo:gradupdate}
    \EndFor
        \State $\psi \leftarrow \psi + \frac{\beta}{m}\sum_{i=1}^{m}(\psi_{i}^{(k)} - \psi)$ \label{algo:reptileupdate}
    \EndWhile
\State \textbf{return} Meta-Adapter parameter $\psi$
\end{algorithmic}
\end{algorithm}

\section{Experiments}
\label{exp}

\noindent \textbf{Datasets.}
We split the datasets in two groups: \textit{meta-training} or \textit{training  dataset} (used in step 1, $\S$ \ref{method:meta_training}) and \textit{meta-adapting} or \textit{adapting dataset} (used in step 2,  $\S$ \ref{method:meta_adapt}).
We first meta-learn the adapters on the training dataset (that contains DLPs different to the ones we will evaluate on), then  fine-tune to new domains and language pairs on the adapting dataset (a small dataset of the DLPs we will evaluate on). 
We list the datasets used, each treated as a different domain: \textit{EUbookshop, KDE, OpenSubtitles, QED, TED, Ubuntu, Bible, UN, Tanzil, Infopankki}. The datasets cover the following languages (ISO 639-1 language code\footnote{\url{https://en.wikipedia.org/wiki/List_of_ISO_639-1_codes}}): \textit{en, de, fr, mk, sr, et, hr, hu, fi, uk, is, lt, ar, es, ru, zh} and are publicly available on OPUS\footnote{\url{https://opus.nlpl.eu}} \cite{tiedemann-2012-parallel}.

\noindent \textbf{Data Preprocessing.}
For each \textit{training dataset}, we strictly limit the corpus of each DLP to a maximum of 5000 sentences to simulate a low-resource setting.
For each \textit{adapting dataset}, we use 500 sentences in each DLP to fine-tune the MNMT model, simulating a few-shot setting.
For the validation and test set, we select 500 sentences and avoid overlap with the adapting dataset by de-duplication.
We filter out sentences longer than 175 tokens and preprocess all data using sentencepiece\footnote{\url{github.com/google/sentencepiece}} \cite{kudo-richardson-2018-sentencepiece}.
More details for the data used in this paper can be found in the Appendix~\ref{appendix:dataset}.

\noindent \textbf{Baselines.}
We compare \md with the following baselines:
i) \textbf{m2m}: Using the original m2m model \cite{fan2021beyond} to generate the translations.
ii) \textbf{m2m + FT}: Fine-tuning m2m on all DLPs.
iii) \textbf{m2m + tag}: Fine-tuning m2m with domain tags \cite{kobus-etal-2017-domain} on all DLPs.
iv) \textbf{agnostic-adapter}: Mixing the data from all DLPs to train the adapters \cite{cooper-stickland-etal-2021-recipes}, to obtain language and domain-agnostic adapters.
v) \textbf{stack-adapter}: Training two adapters for each language pair and domain, then stacking both adapters \cite{cooper-stickland-etal-2021-multilingual}. Taking `Ubuntu-en-sr' as an example, this approach first trains a \textit{language pair adapter} for `en-sr' using all data containing `en-sr' in all domains (also including the `Ubuntu' domain) and a  \textit{domain adapter} for `Ubuntu' using all data covering all language pairs in the  `Ubuntu' domain. Then, the two adapters are stacked together.
vi) \textbf{meta-learning}: Traditional meta-learning methods using the MAML algorithm \cite{sharaf-etal-2020-meta} on all DLPs.

\noindent \textbf{Implementation.}
We use m2m, released in the HuggingFace repository\footnote{\url{github.com/huggingface/transformers}} \cite{wolf-etal-2020-transformers}.
For adapter training, we use the implementation of the AdapterHub repository\footnote{\url{github.com/adapter-hub/adapter-transformers}} \cite{pfeiffer-etal-2020-adapterhub}.
We use DeepSpeed\footnote{\url{github.com/microsoft/DeepSpeed}} \cite{rasley2020deepspeed} to accelerate the pre-training of big models.
Note that all baseline systems except \textit{stack-adapter} train a single MNMT model or a single adapter on all DLPs in the \textit{training datasets} and then fine-tune to a specific DLP on a single \textit{adapting dataset}.
For \textit{stack-adapter}, the number of language pair adapters and domain adapters to be trained is proportional to the number of language pairs and the number of domains (see more details in Appendix~\ref{appendix:model_config}).

\noindent \textbf{Evaluation.}
We measure case-sensitive detokenized BLEU with SacreBLEU\footnote{\url{github.com/mjpost/sacrebleu}} \cite{post-2018-call}.
For Chinese we use the SacreBLEU tokenizer (\emph{tok zh}) and convert all traditional characters generated by the model to simplified characters using HanziConv.\footnote{\url{github.com/berniey/hanziconv}}
We also evaluate our models using chrF \cite{popovic-2015-chrf} due to the recent criticism of BLEU score \cite{mathur-etal-2020-tangled}; the results are listed in the Appendix~\ref{appendix:chrf_eva}.

\begin{table}[t]
\resizebox{\columnwidth}{!}{
\begin{tabular}{lcccc}
\toprule
\multirow{2}{*}{} & \multirow{2}{*}{\textbf{BLEU}} & \multicolumn{3}{c}{\textbf{specific domain}} \\
\cmidrule(lr){3-5}
 &  & TED & Ubuntu & KDE \\
\hline
\textbf{m2m} & 18.18 & 16.20 & 20.61 & 22.04 \\
\textbf{m2m + FT} & 20.84 & 17.53 & 28.81 & 29.19 \\
\textbf{m2m + tag} & 22.70 & 18.70 & \textbf{31.86} & 31.53 \\
\textbf{agnostic-adapter} & 23.70 & \textbf{19.82} & 31.07 & 32.74 \\
\textbf{stack-adapter} & 21.06 & 18.34 & 29.17 & 30.26 \\
\textbf{meta-learning} & 20.01 & 17.57 & 28.11 & 28.59 \\
\boldmath \md & \textbf{23.89} & 19.77 & 31.46 & \textbf{32.91} \\
\bottomrule
\end{tabular}}
\caption{\label{tab:res_seen}
Performance on the \textit{meta-training stage} (DLPs of \textit{training dataset}): average BLEU on DLPs over all domains (left); average BLEU on DLPs per domain (right, under \textit{specific domain}).}
\end{table}

\begin{table*}[t]
\resizebox{\textwidth}{!}{
\begin{tabular}{lccc||ccc||ccc}
\hline
 & \multicolumn{3}{c}{\textbf{DLP (meta-adaptation  domain)}} & \multicolumn{6}{c}{\textbf{specific DLP}} \\
 \hline
 & UN & Tanzil & Infopankki & UN-ar-en & Tanzil-ar-en & Infopankki-ar-en & UN-ar-ru & Tanzil-ar-ru & Infopankki-ar-ru \\
\hline 
\textbf{m2m} & 32.28 & 8.72 & 17.40 & 38.94 & 6.44 & 22.57 & 22.96 & 3.64 & 15.05 \\
\textbf{m2m + FT} & 29.93 & 8.26 & 15.88 & 35.11 & 6.85 & 21.33 & 19.10 & 3.05 & 14.19 \\
\textbf{m2m + tag} & 29.88 & 8.06 & 15.93 & 34.39 & 6.63 & 20.12 & 19.37 & 2.65 & 13.68 \\
\textbf{agnostic-adapter} & 30.56 & 8.42 & 17.36 & 36.13 & 6.12 & 23.08 & 20.64 & 3.63 & 14.96 \\
\textbf{stack-adapter} & 29.64 & 8.14 & 17.19 & 35.31 & 5.83 & 22.14 & 19.17 & 2.34 & 13.85 \\
\textbf{meta-learning} & 32.21 & 7.02 & 16.73 & 37.13 & 5.50 & 18.91 & 22.68 & 1.70 & 15.23 \\
\hline
\textbf{\boldmath \md} & \textbf{33.53} & \textbf{9.87} & \textbf{18.43} & \textbf{39.05} & \textbf{8.56} & \textbf{23.21} & \textbf{25.22} & \textbf{4.33} & \textbf{17.48} \\
\boldmath $\Delta$ & +1.25 & +1.15 & +1.03 & +0.11 & +2.12 & +0.64 & +2.26 & +0.69 & +2.43 \\
\hline
\end{tabular}}
\caption{\label{tab:main_res}
Main results on the \textit{meta-adaptation stage}: average BLEU scores on all DLPs with different adaptation domain (left) and BLEU scores on some examples of specific DLP (right). $\Delta$ denotes improvement over m2m.}
\end{table*}

\begin{table}[t]
\centering
\resizebox{1\columnwidth}{!}{
\begin{tabular}{lrrr}
\toprule
\textbf{Method} & \textbf{\#Param.} & \textbf{Time$_T$} & \textbf{Time$_A$}\\
\hline
\textbf{m2m} & 418M (100\%) & - & - \\
\textbf{m2m + FT} & 418M (100\%) & 100\% & 100\% \\
\textbf{m2m + tag}  & 418M (100\%) & 100\% & 100\% \\
\textbf{agnostic-adapter} & 3.17M (0.75\%) & 42\% & 150\% \\
\textbf{stack-adapter}  & $k\cdot$ 3.17M ($k\cdot$ 0.75\%) & $k\cdot$ 42\% & 200\%  \\
\textbf{meta-learning} & 418M (100\%) & 75\% & 500\% \\
\boldmath \md & 3.17M (0.75\%) & 34\% & 300\% \\
\bottomrule
\end{tabular}}
\caption{Number of trainable parameters and Training/Adapting time relative to fine-tuning. $k$ denotes the number of DLPs during the training process.} 
\label{tab:efficiency}
\end{table}

\begin{table*}[t]
\resizebox{\textwidth}{!}{
\begin{tabular}{lccccccc||ccccccc}
\hline
 & \multicolumn{7}{c}{\textbf{meta-adaptation domain}} & \multicolumn{7}{c}{\textbf{specific DLP (hr-sr)}} \\
\hline 
 & \multicolumn{1}{l}{EUbookshop} & \multicolumn{1}{l}{KDE} & \multicolumn{1}{l}{OpenSubtitles} & \multicolumn{1}{l}{QED} & \multicolumn{1}{l}{TED} & \multicolumn{1}{l}{Ubuntu} & \multicolumn{1}{l}{Bible} & EUbookshop & KDE & OpenSubtitles & QED & TED & Ubuntu & Bible \\
\hline 
\textbf{m2m} & 17.77 & 22.05 & 14.13 & 18.34 & 16.20 & 20.62 & 9.80 & 11.43 & 25.37 & 19.01 & 12.25 & 8.14 & 22.33 & 2.01 \\
\textbf{m2m + FT} & 12.73 & 24.56 & 16.22 & 20.46 & 18.74 & 31.32 & 11.30 & 9.79 & 21.05 & 53.34 & 23.87 & 20.81 & \textbf{34.08} & 12.57 \\
\textbf{m2m + tag} & 13.03 & 25.34 & 16.12 & 17.75 & 17.04 & 26.29 & 11.49 & 10.13 & 29.64 & 49.54 & 19.78 & 20.43 & 34.15 & 13.25 \\
\textbf{agnostic-adapter} & 16.24 & \textbf{25.85} & 17.90 & 21.71 & \textbf{20.08} & \textbf{31.53} & 11.75 & 9.05 & \textbf{30.64} & 54.04 & 22.79 & 21.19 & 28.83 & 10.59 \\
\textbf{stack-adapter} & 13.25 & 24.19 & 17.21 & 19.56 & 18.37 & 28.27 & 10.38 & 10.55 & 24.50 & 42.94 & 22.02 & 20.95 & 25.41 & 10.14 \\
\textbf{meta-learning} & 13.61 & 24.91 & 16.22 & 17.70 & 16.40 & 24.93 & 11.84 & 7.90 & 27.85 & 52.50 & 20.41 & 19.00 & 31.24 & 10.42 \\
\hline
\textbf{\boldmath \md} & \textbf{18.99} & 25.22 & \textbf{17.94} & \textbf{21.71} & 19.86 & 31.37 & \textbf{12.12} & \textbf{12.05} & 30.49 & \textbf{54.30} & \textbf{23.92} & \textbf{21.32} & 33.71 & \textbf{13.69} \\
\boldmath $\Delta$ & +2.75 & -0.63 & +0.04 & +0.00 & -0.22 & -0.16 & +0.37 & +3.00 & -0.15 & +0.26 & +1.13 & +0.13 & +4.88 & +3.1 \\
\hline
\end{tabular}}
\caption{Domain transfer via languages: average BLEU scores on all DLPs in each \textit{meta-adaptation} domain (left) and BLEU scores on a random selection of one specific DLP in \textit{hr-sr} (right). $\Delta$ denotes improvement over  \textit{agnostic-adapter}.}
\label{tab:domain_transfer}
\end{table*}

\section{Results}
\label{sec:res}
Our goal is to evaluate the adaptability of \md on a variety of new domains and new language pairs simultaneously.
In the \textit{meta-training} stage, we perform meta-learning of the model on 180 DLPs, which contain 6 domains (\textit{EUbookshop, KDE, OpenSubtitles, QED, TED, Ubuntu}) and 30 language pairs (\textit{en, et, mk, sr, hr, hu}), comparing our approach to different baseline systems.
In the \textit{meta-adaptation} stage, we fine-tune both our model and the baselines to 3 domains (\textit{UN, Tanzil, Infopankki}) and 30 language pairs (using \textit{ar, en, es, fr, ru, zh}) of the same dataset simultaneously.
Table~\ref{tab:res_seen} shows the results in the \textit{meta-training} step and Table~\ref{tab:main_res} presents the main results of our model in the \textit{meta-adaptation} step compared to the baselines (results for all DLPs are in Appendix~\ref{appendix:all_dlps}).

Motivated by \citet{lai-etal-2022-improving-domain}, we compare our approach to multiple baselines in terms of domain robustness.
As shown in Table~\ref{tab:res_seen}, \md obtains a performance that is on par or better than \textit{agnostic-adapter}, which is a robust model.
Note that \md also outperforms \textit{m2m+tag}, which was shown to be the most robust model in \citet{cooper-stickland-etal-2021-multilingual}.
After showing empirically that we obtain a robust model, we verify its adaptability (see Table~\ref{tab:main_res} and  $\S$~\ref{analysis:dtvl}) and language transfer ability ($\S$~\ref{analysis:ltvd}) through a series of experiments.

As shown in Table~\ref{tab:main_res}, \md performs well when adapting to the \textit{meta-adaptation} domains and language pairs at the same time.
We observe that no baseline system outperforms the original m2m model. This implies that these models are unable to transfer language or domain knowledge from the MNMT model. One possible explanation is that these models already exhibit over-fitting and catastrophic forgetting when trained on \textit{meta-training} domains and language pairs in such limited resource scenarios.

Because of the unpredictability of the baseline systems' performance, it is difficult to draw reliable conclusions. For example, in the UN domain, \textit{meta-learning} is on par with the original m2m model. However, performance on Tanzil and Infopankki is much worse than the one of the original m2m model. The \textit{agnostic-adapter} also performs comparably with the original m2m model in the same domains, which shows that it is a robust model. Still, it obtains much worse performance on UN. In contrast, \md has a more stable performance when adapting to new domains and language pairs. 

In addition, \md has the ability to improve the performance of some DLPs on which baseline models obtain extremely low BLEU scores, especially in some distant domains. For example, in Tanzil-ar-ru, the traditional meta-learning method only gets 1.70 BLEU score, while \md gets 4.33.

\section{Analysis}
In this section, we conduct additional experiments to better understand the strengths of \md.
We first investigate the benefits in terms of speed in \md training and adapting (Section~\ref{analysis:efficiency}), then investigate the cross-lingual domain transfer and cross-domain language transfer through an ablation study  (Section~\ref{analysis:ablation}).

\subsection{Efficiency of \boldmath \md}
\label{analysis:efficiency}
We compare the efficiency of baselines to traditional fine-tuning and list their number of trainable parameters and training/adapting time in Table~\ref{tab:efficiency}.

\md only updates the adapter parameters while freezing the MNMT model's parameters (just like \textit{agnostic-adapter}). Therefore, it has fewer trainable parameters compared to fine-tuning (0.75\% of the parameters of the entire model).
Furthermore, the parameters of \md are significantly fewer than those of \textit{stack-adapter}, which are $k$ times larger than those of standard adapter-based approaches. This happens because domain adapters and language pair adapters must be trained in each DLP when training the \textit{stack-adapter} model.
Adapter-based approaches train $34\%$-$42\%$ faster than fine-tuning due to parameter efficiency.
The adaptation time of \md, on the other hand, is often longer since it requires updating the high-level gradient. 
Our approach requires more time than traditional adapter methods but is faster compared with updating
the entire model using traditonal meta-learning.
For example, the adaptation time for \textit{m2m+FT} is 40$s$, while for \md it is 120$s$, which is still a lot faster than standard \textit{meta-learning} (200$s$).

\begin{table*}[t]
\resizebox{\textwidth}{!}{
\begin{tabular}{lcccc||cccccc}
\hline
 & \multicolumn{4}{c}{\textbf{meta-adaptation language pair}} & \multicolumn{6}{c}{\textbf{specific DLP (de-en)}} \\
\hline
 & de-en & en-fr & fi-uk & is-lt & EUbookshop & KDE & OpenSubtitles & QED & TED & Ubuntu \\
\hline
\textbf{m2m} & 24.52 & 29.20 & 12.34 & 12.55 & 19.59 & 26.48 & 15.89 & 26.34 & 28.14 & 30.65 \\
\textbf{m2m + FT} & 23.29 & 24.44 & 11.29 & 9.59 & 16.04 & 23.17 & 13.34 & 21.39 & 26.20 & 39.59 \\
\textbf{m2m + tag} & 22.52 & 24.97 & 11.71 & 11.22 & 15.86 & 23.67 & 11.72 & 20.64 & 25.97 & 37.25 \\
\textbf{agnostic-adapter} & 28.33 & \textbf{30.93} & \textbf{15.42} & \textbf{14.38} & 20.16 & \textbf{28.72} & 17.97 & \textbf{27.66} & \textbf{33.63} & 41.89 \\
\textbf{stack-adapter} & 23.37 & 24.96 & 11.51 & 11.09 & 16.14 & 22.51 & 13.84 & 22.29 & 27.67 & 36.73 \\
\textbf{meta-learning} & 25.08 & 28.26 & 13.40 & 12.83 & 17.88 & 21.20 & 16.32 & 24.96 & 30.32 & 39.81 \\
\hline
\boldmath \md & \textbf{28.37} & 30.80 & 15.24 & 14.05 & \textbf{20.20} & 28.19 & \textbf{18.06} & 27.18 & 33.32 & \textbf{43.24} \\
$\Delta$ & +0.04 & -0.13 & -0.18 & -0.33 & +0.04 & -0.53 & +0.09 & -0.48 & -0.31 & +1.35 \\
\hline
\end{tabular}}
\caption{Language transfer via domains: average BLEU scores on all DLPs in each meta-adaptation language pair (left) and BLEU scores on one specific DLP in \textit{de-en} (right). $\Delta$ denotes the improvement over the  \textit{agnostic-adapter}.}
\label{tab:lang_transfer}
\end{table*}

\subsection{Ablation Study}
\label{analysis:ablation}
We conduct a number of experiments with extensive analysis to validate the domain transfer ability of the \md across different language pairs ($\S$~\ref{analysis:dtvl}), as well as the language transfer ability across multiple domains ($\S$~\ref{analysis:ltvd}). 

\subsubsection{Domain Transfer via Languages}
\label{analysis:dtvl}
To investigate the capacity of our models to transfer domain knowledge across different languages, we define domain transfer via languages, i.e., the ability to transfer domains while keeping the languages unchanged.
We first fine-tune the MNMT model in some of the \textit{meta-training} domains under the specified language pair, and then we adapt these trained models to new \textit{meta-adaptation} domains of the same language pair.
To be more specific, we first choose 6 languages (\textit{en, et, mk, sr, hr, hu}), forming 30 language pairs. 
Then, we choose six of these seven domains (\textit{EUbookshop, KDE, OpenSubtitles, QED, TED, Ubuntu, Bible}) across all selected 30 language pairs as the \textit{meta-training} dataset (180 DLPs) to fine-tune the MNMT model, and another one domain as the adapting domain across all selected language pairs (30 DLPs) to evaluate the adaptability of the
fine-tuned MNMT model to the new domain.
Table~\ref{tab:domain_transfer} provides the results for domain transfer across languages.

From Table~\ref{tab:domain_transfer}, we observe that almost all baseline systems and \md outperform the original m2m model (except for the \textit{EUbookshop} domain), indicating that the model encodes language knowledge and can transfer this knowledge to new \textit{meta-adaptation} domains.
Our approach is comparable to the performance of \textit{agnostic-adapter}, which performs the best among all baseline systems.

We also discover that domain transfer through languages is desirable in some distant domains. For example, the original m2m model only got BLEU scores of 2.01 and 19.01 in the \textit{Bible} and \textit{OpenSubtitles} domain (\textit{hr-sr} language pair). However, domain transfer through \md resulted in a considerable performance boost and achieved a BLEU score of 13.69 and 54.30.

We notice that none of the baselines outperforms the original m2m model in the \textit{EUbookshop} domain, which means that the language knowledge learned from the baseline model does not transfer to this particular domain.
Our approach, on the other hand, has a strong domain transfer ability.
We investigated the reason, which was caused by a significant overfitting issue while adapting to the \textit{EUbookshop} domain. The previous fine-tuning strategy converged too early, resulting in significant overfitting of the model to the \textit{meta-training} dataset, which performed exceedingly badly in adapting to the new domain (see the loss decline curve in Appendix~\ref{appendix:res_EU} for more details).
This phenomenon is also consistent with our previous findings ($\S$~\ref{sec:res}) that our approach is more stable than the baseline systems in adapting to new domains.

\subsubsection{Language Transfer via Domains}
\label{analysis:ltvd}
To study the ability of our model to transfer language knowledge across different domains, we define language transfer via domains, i.e., the ability to transfer languages while keeping the domains unchanged.
To this end, we first fine-tune the MNMT model in some \textit{meta-training} DLPs, and then we adapt these trained models to \textit{meta-adaptation} language pairs of the same domains.
To achieve this, we first select 180 DLPs as the \textit{meta-training} dataset to train the model, which contains 6 domains (\textit{EUbookshop, KDE, OpenSubtitles, QED, TED, Ubuntu}) and 30 language pairs (\textit{en, et, mk, sr, hr, hu}); then adapt these trained model to 4 of the \textit{meta-adaptation} language pairs (\textit{de-en, en-fr, fi-uk, is-lt}).
The findings of language transfer across domains are shown in Table~\ref{tab:lang_transfer}.

According to Table~\ref{tab:lang_transfer}, the performance of traditional fine-tuning approaches (\textit{m2m+FT}, \textit{m2m+tag}) are poorer than the original m2m model, which means that these methods do not transfer the learned domain knowledge to the new \textit{meta-adaptation} language pair.
This meets our expectation since m2m is trained on a big dataset and learns a great quantity of linguistic information, which limits its capacity to transfer language information in small datasets.
This explanation can be demonstrated by the results of the \textit{meta-learning} approach. As shown in Table~\ref{tab:lang_transfer},  \textit{meta-learning} yields slightly higher BLEU scores compared to the original m2m model, which arguably supports the conclusion that the original m2m model already has strong linguistic information. These small improvements from \textit{meta-learning} can be attributed to leveraging the limited data available.

In contrast, adapter-based methods (\textit{agnostic-adapter} and \md) permit cross-lingual transfer across domains.
\md shows a performance that is on par or better than the \textit{agnostic-adapter}, the most competitive model in all baseline systems.
The results of the \textit{stack-adapter} show that it cannot perform language transfer across domains through naively stacking domain adapters and language adapters. This is consistent with the conclusions of \citet{cooper-stickland-etal-2021-multilingual}.

Similarly, \md has demonstrated significant language transfer ability in distant domains. In \textit{ubuntu-de-en}, for example, \md achieves a BLEU score of 43.24, which is significantly higher than the original m2m model's BLEU of 30.65.

\section{Conclusion}
We present \md, a novel multilingual multi-domain NMT adaptation framework which combines meta-learning and parameter-efficient fine-tuning with adapters.
\md is effective on adapting to new languages and domains simultaneously in low-resource settings.
We find that \md also transfers language knowledge across domains and transfers domain information across languages.
In addition, \md is efficient in training and adaptation, which is practical for online adaptation \cite{etchegoyhen-etal-2021-online} to complex scenarios (new languages and new domains) in the real world.

\section{Limitations}
This work has two main limitations.
i) We have only evaluated the proposed method on limited and balanced bilingual training data to simulate the low-resource scenario. However, some domains in our setting are in fact highly imbalanced.
ii) We only evaluated \md on machine translation, perhaps it would be plausible to expand our method to other NLP tasks, such as text generation or language modeling. Since our framework leverages a multilingual pretrained model and only trains adapters, we believe it could easily be applied to other tasks besides MT.

\section*{Acknowledgement}
This work was supported by funding to Wen Lai's PhD research from LMU-CSC (China Scholarship Council) Scholarship Program.
This work has received funding from the European Research Council under the European Union’s Horizon $2020$ research and innovation program (grant agreement 
\#$640550$). This work was also supported by the DFG (grant FR $2829$/$4$-$1$). 

\bibliography{anthology,custom}

\begin{thebibliography}{50}
\expandafter\ifx\csname natexlab\endcsname\relax\def\natexlab#1{#1}\fi

\bibitem[{Aharoni and Goldberg(2020)}]{aharoni-goldberg-2020-unsupervised}
Roee Aharoni and Yoav Goldberg. 2020.
\newblock \href {https://doi.org/10.18653/v1/2020.acl-main.692} {Unsupervised
  domain clusters in pretrained language models}.
\newblock In \emph{Proceedings of the 58th Annual Meeting of the Association
  for Computational Linguistics}, pages 7747--7763, Online. Association for
  Computational Linguistics.

\bibitem[{Aharoni et~al.(2019)Aharoni, Johnson, and
  Firat}]{aharoni-etal-2019-massively}
Roee Aharoni, Melvin Johnson, and Orhan Firat. 2019.
\newblock \href {https://doi.org/10.18653/v1/N19-1388} {Massively multilingual
  neural machine translation}.
\newblock In \emph{Proceedings of the 2019 Conference of the North {A}merican
  Chapter of the Association for Computational Linguistics: Human Language
  Technologies, Volume 1 (Long and Short Papers)}, pages 3874--3884,
  Minneapolis, Minnesota. Association for Computational Linguistics.

\bibitem[{Ba et~al.(2016)Ba, Kiros, and Hinton}]{ba2016layer}
Jimmy~Lei Ba, Jamie~Ryan Kiros, and Geoffrey~E Hinton. 2016.
\newblock \href {https://arxiv.org/abs/1607.06450} {Layer normalization}.
\newblock \emph{arXiv preprint arXiv:1607.06450}.

\bibitem[{Bapna and Firat(2019)}]{bapna-firat-2019-simple}
Ankur Bapna and Orhan Firat. 2019.
\newblock \href {https://doi.org/10.18653/v1/D19-1165} {Simple, scalable
  adaptation for neural machine translation}.
\newblock In \emph{Proceedings of the 2019 Conference on Empirical Methods in
  Natural Language Processing and the 9th International Joint Conference on
  Natural Language Processing (EMNLP-IJCNLP)}, pages 1538--1548, Hong Kong,
  China. Association for Computational Linguistics.

\bibitem[{Chen et~al.(2022)Chen, Zhong, Zha, Karypis, and
  He}]{chen-etal-2022-meta}
Yanda Chen, Ruiqi Zhong, Sheng Zha, George Karypis, and He~He. 2022.
\newblock \href {https://doi.org/10.18653/v1/2022.acl-long.53} {Meta-learning
  via language model in-context tuning}.
\newblock In \emph{Proceedings of the 60th Annual Meeting of the Association
  for Computational Linguistics (Volume 1: Long Papers)}, pages 719--730,
  Dublin, Ireland. Association for Computational Linguistics.

\bibitem[{Chronopoulou et~al.(2022)Chronopoulou, Stojanovski, and
  Fraser}]{chronopoulou2022language}
Alexandra Chronopoulou, Dario Stojanovski, and Alexander Fraser. 2022.
\newblock \href {https://arxiv.org/abs/2209.15236} {Language-family adapters
  for multilingual neural machine translation}.
\newblock \emph{arXiv preprint arXiv:2209.15236}.

\bibitem[{Chu and Dabre(2019)}]{chu2019multilingual}
Chenhui Chu and Raj Dabre. 2019.
\newblock \href {https://arxiv.org/abs/1906.07978} {Multilingual multi-domain
  adaptation approaches for neural machine translation}.
\newblock \emph{arXiv preprint arXiv:1906.07978}.

\bibitem[{Chu and Wang(2018)}]{chu-wang-2018-survey}
Chenhui Chu and Rui Wang. 2018.
\newblock \href {https://aclanthology.org/C18-1111} {A survey of domain
  adaptation for neural machine translation}.
\newblock In \emph{Proceedings of the 27th International Conference on
  Computational Linguistics}, pages 1304--1319, Santa Fe, New Mexico, USA.
  Association for Computational Linguistics.

\bibitem[{Cooper~Stickland et~al.(2021{\natexlab{a}})Cooper~Stickland, Berard,
  and Nikoulina}]{cooper-stickland-etal-2021-multilingual}
Asa Cooper~Stickland, Alexandre Berard, and Vassilina Nikoulina.
  2021{\natexlab{a}}.
\newblock \href {https://aclanthology.org/2021.wmt-1.64} {Multilingual domain
  adaptation for {NMT}: Decoupling language and domain information with
  adapters}.
\newblock In \emph{Proceedings of the Sixth Conference on Machine Translation},
  pages 578--598, Online. Association for Computational Linguistics.

\bibitem[{Cooper~Stickland et~al.(2021{\natexlab{b}})Cooper~Stickland, Li, and
  Ghazvininejad}]{cooper-stickland-etal-2021-recipes}
Asa Cooper~Stickland, Xian Li, and Marjan Ghazvininejad. 2021{\natexlab{b}}.
\newblock \href {https://doi.org/10.18653/v1/2021.eacl-main.301} {Recipes for
  adapting pre-trained monolingual and multilingual models to machine
  translation}.
\newblock In \emph{Proceedings of the 16th Conference of the European Chapter
  of the Association for Computational Linguistics: Main Volume}, pages
  3440--3453, Online. Association for Computational Linguistics.

\bibitem[{Dakwale and Monz(2017)}]{dakwale2017finetuning}
Praveen Dakwale and Christof Monz. 2017.
\newblock \href
  {https://staff.science.uva.nl/c.monz/ltl/publications/mtsummit2017.pdf}
  {Finetuning for neural machine translation with limited degradation across
  in-and out-of-domain data}.
\newblock \emph{Proceedings of the XVI Machine Translation Summit}, 117.

\bibitem[{Domhan and Hieber(2017)}]{domhan-hieber-2017-using}
Tobias Domhan and Felix Hieber. 2017.
\newblock \href {https://doi.org/10.18653/v1/D17-1158} {Using target-side
  monolingual data for neural machine translation through multi-task learning}.
\newblock In \emph{Proceedings of the 2017 Conference on Empirical Methods in
  Natural Language Processing}, pages 1500--1505, Copenhagen, Denmark.
  Association for Computational Linguistics.

\bibitem[{Dou et~al.(2019)Dou, Yu, and
  Anastasopoulos}]{dou-etal-2019-investigating}
Zi-Yi Dou, Keyi Yu, and Antonios Anastasopoulos. 2019.
\newblock \href {https://doi.org/10.18653/v1/D19-1112} {Investigating
  meta-learning algorithms for low-resource natural language understanding
  tasks}.
\newblock In \emph{Proceedings of the 2019 Conference on Empirical Methods in
  Natural Language Processing and the 9th International Joint Conference on
  Natural Language Processing (EMNLP-IJCNLP)}, pages 1192--1197, Hong Kong,
  China. Association for Computational Linguistics.

\bibitem[{Etchegoyhen et~al.(2021)Etchegoyhen, Ponce, Gete, and
  Ruiz}]{etchegoyhen-etal-2021-online}
Thierry Etchegoyhen, David Ponce, Harritxu Gete, and Victor Ruiz. 2021.
\newblock \href {https://aclanthology.org/2021.ranlp-1.47} {Online learning
  over time in adaptive neural machine translation}.
\newblock In \emph{Proceedings of the International Conference on Recent
  Advances in Natural Language Processing (RANLP 2021)}, pages 411--420, Held
  Online. INCOMA Ltd.

\bibitem[{Fan et~al.(2021)Fan, Bhosale, Schwenk, Ma, El-Kishky, Goyal, Baines,
  Celebi, Wenzek, Chaudhary et~al.}]{fan2021beyond}
Angela Fan, Shruti Bhosale, Holger Schwenk, Zhiyi Ma, Ahmed El-Kishky,
  Siddharth Goyal, Mandeep Baines, Onur Celebi, Guillaume Wenzek, Vishrav
  Chaudhary, et~al. 2021.
\newblock \href {https://arxiv.org/abs/2010.11125} {Beyond english-centric
  multilingual machine translation}.
\newblock \emph{Journal of Machine Learning Research}, 22(107):1--48.

\bibitem[{Finn et~al.(2017)Finn, Abbeel, and Levine}]{finn2017model}
Chelsea Finn, Pieter Abbeel, and Sergey Levine. 2017.
\newblock \href {https://proceedings.mlr.press/v70/finn17a.html}
  {Model-agnostic meta-learning for fast adaptation of deep networks}.
\newblock In \emph{International conference on machine learning}, pages
  1126--1135. PMLR.

\bibitem[{Freitag and Al-Onaizan(2016)}]{freitag2016fast}
Markus Freitag and Yaser Al-Onaizan. 2016.
\newblock \href {https://arxiv.org/abs/1612.06897} {Fast domain adaptation for
  neural machine translation}.
\newblock \emph{arXiv preprint arXiv:1612.06897}.

\bibitem[{Gu et~al.(2018)Gu, Wang, Chen, Li, and Cho}]{gu-etal-2018-meta}
Jiatao Gu, Yong Wang, Yun Chen, Victor O.~K. Li, and Kyunghyun Cho. 2018.
\newblock \href {https://doi.org/10.18653/v1/D18-1398} {Meta-learning for
  low-resource neural machine translation}.
\newblock In \emph{Proceedings of the 2018 Conference on Empirical Methods in
  Natural Language Processing}, pages 3622--3631, Brussels, Belgium.
  Association for Computational Linguistics.

\bibitem[{Gu et~al.(2019)Gu, Wang, Cho, and Li}]{gu-etal-2019-improved}
Jiatao Gu, Yong Wang, Kyunghyun Cho, and Victor~O.K. Li. 2019.
\newblock \href {https://doi.org/10.18653/v1/P19-1121} {Improved zero-shot
  neural machine translation via ignoring spurious correlations}.
\newblock In \emph{Proceedings of the 57th Annual Meeting of the Association
  for Computational Linguistics}, pages 1258--1268, Florence, Italy.
  Association for Computational Linguistics.

\bibitem[{Houlsby et~al.(2019)Houlsby, Giurgiu, Jastrzebski, Morrone,
  De~Laroussilhe, Gesmundo, Attariyan, and Gelly}]{houlsby2019parameter}
Neil Houlsby, Andrei Giurgiu, Stanislaw Jastrzebski, Bruna Morrone, Quentin
  De~Laroussilhe, Andrea Gesmundo, Mona Attariyan, and Sylvain Gelly. 2019.
\newblock \href {http://proceedings.mlr.press/v97/houlsby19a.html}
  {Parameter-efficient transfer learning for nlp}.
\newblock In \emph{International Conference on Machine Learning}, pages
  2790--2799. PMLR.

\bibitem[{Johnson et~al.(2017)Johnson, Schuster, Le, Krikun, Wu, Chen, Thorat,
  Vi{\'e}gas, Wattenberg, Corrado, Hughes, and
  Dean}]{johnson-etal-2017-googles}
Melvin Johnson, Mike Schuster, Quoc~V. Le, Maxim Krikun, Yonghui Wu, Zhifeng
  Chen, Nikhil Thorat, Fernanda Vi{\'e}gas, Martin Wattenberg, Greg Corrado,
  Macduff Hughes, and Jeffrey Dean. 2017.
\newblock \href {https://doi.org/10.1162/tacl_a_00065} {{G}oogle{'}s
  multilingual neural machine translation system: Enabling zero-shot
  translation}.
\newblock \emph{Transactions of the Association for Computational Linguistics},
  5:339--351.

\bibitem[{Khayrallah et~al.(2017)Khayrallah, Kumar, Duh, Post, and
  Koehn}]{khayrallah-etal-2017-neural}
Huda Khayrallah, Gaurav Kumar, Kevin Duh, Matt Post, and Philipp Koehn. 2017.
\newblock \href {https://aclanthology.org/I17-2004} {Neural lattice search for
  domain adaptation in machine translation}.
\newblock In \emph{Proceedings of the Eighth International Joint Conference on
  Natural Language Processing (Volume 2: Short Papers)}, pages 20--25, Taipei,
  Taiwan. Asian Federation of Natural Language Processing.

\bibitem[{Kobus et~al.(2017)Kobus, Crego, and
  Senellart}]{kobus-etal-2017-domain}
Catherine Kobus, Josep Crego, and Jean Senellart. 2017.
\newblock \href {https://doi.org/10.26615/978-954-452-049-6_049} {Domain
  control for neural machine translation}.
\newblock In \emph{Proceedings of the International Conference Recent Advances
  in Natural Language Processing, {RANLP} 2017}, pages 372--378, Varna,
  Bulgaria. INCOMA Ltd.

\bibitem[{Kudo and Richardson(2018)}]{kudo-richardson-2018-sentencepiece}
Taku Kudo and John Richardson. 2018.
\newblock \href {https://doi.org/10.18653/v1/D18-2012} {{S}entence{P}iece: A
  simple and language independent subword tokenizer and detokenizer for neural
  text processing}.
\newblock In \emph{Proceedings of the 2018 Conference on Empirical Methods in
  Natural Language Processing: System Demonstrations}, pages 66--71, Brussels,
  Belgium. Association for Computational Linguistics.

\bibitem[{Lai et~al.(2022)Lai, Libovick{\'y}, and
  Fraser}]{lai-etal-2022-improving-domain}
Wen Lai, Jind{\v{r}}ich Libovick{\'y}, and Alexander Fraser. 2022.
\newblock \href {https://aclanthology.org/2022.coling-1.461} {Improving both
  domain robustness and domain adaptability in machine translation}.
\newblock In \emph{Proceedings of the 29th International Conference on
  Computational Linguistics}, pages 5191--5204, Gyeongju, Republic of Korea.
  International Committee on Computational Linguistics.

\bibitem[{Lee et~al.(2022)Lee, Li, and Vu}]{lee2022meta}
Hung-yi Lee, Shang-Wen Li, and Ngoc~Thang Vu. 2022.
\newblock \href {https://arxiv.org/abs/2205.01500} {Meta learning for natural
  language processing: A survey}.
\newblock \emph{arXiv preprint arXiv:2205.01500}.

\bibitem[{Loshchilov and Hutter(2018)}]{loshchilov2018decoupled}
Ilya Loshchilov and Frank Hutter. 2018.
\newblock \href {https://openreview.net/forum?id=Bkg6RiCqY7} {Decoupled weight
  decay regularization}.
\newblock In \emph{International Conference on Learning Representations}.

\bibitem[{Mathur et~al.(2020)Mathur, Baldwin, and
  Cohn}]{mathur-etal-2020-tangled}
Nitika Mathur, Timothy Baldwin, and Trevor Cohn. 2020.
\newblock \href {https://doi.org/10.18653/v1/2020.acl-main.448} {Tangled up in
  {BLEU}: Reevaluating the evaluation of automatic machine translation
  evaluation metrics}.
\newblock In \emph{Proceedings of the 58th Annual Meeting of the Association
  for Computational Linguistics}, pages 4984--4997, Online. Association for
  Computational Linguistics.

\bibitem[{McCloskey and Cohen(1989)}]{mccloskey1989catastrophic}
Michael McCloskey and Neal~J Cohen. 1989.
\newblock \href
  {https://www.sciencedirect.com/science/article/abs/pii/S0079742108605368}
  {Catastrophic interference in connectionist networks: The sequential learning
  problem}.
\newblock In \emph{Psychology of learning and motivation}, volume~24, pages
  109--165. Elsevier.

\bibitem[{Neubig and Hu(2018)}]{neubig-hu-2018-rapid}
Graham Neubig and Junjie Hu. 2018.
\newblock \href {https://doi.org/10.18653/v1/D18-1103} {Rapid adaptation of
  neural machine translation to new languages}.
\newblock In \emph{Proceedings of the 2018 Conference on Empirical Methods in
  Natural Language Processing}, pages 875--880, Brussels, Belgium. Association
  for Computational Linguistics.

\bibitem[{Nichol et~al.(2018)Nichol, Achiam, and Schulman}]{nichol2018first}
Alex Nichol, Joshua Achiam, and John Schulman. 2018.
\newblock \href {https://arxiv.org/abs/1803.02999} {On first-order
  meta-learning algorithms}.
\newblock \emph{arXiv preprint arXiv:1803.02999}.

\bibitem[{Park et~al.(2022)Park, Kim, Calapodescu, Cho, and
  Nikoulina}]{park-etal-2022-dalc}
Cheonbok Park, Hantae Kim, Ioan Calapodescu, Hyun~Chang Cho, and Vassilina
  Nikoulina. 2022.
\newblock \href {https://aclanthology.org/2022.findings-acl.141} {{D}a{LC}:
  Domain adaptation learning curve prediction for neural machine translation}.
\newblock In \emph{Findings of the Association for Computational Linguistics:
  ACL 2022}, pages 1789--1807, Dublin, Ireland. Association for Computational
  Linguistics.

\bibitem[{Park et~al.(2017)Park, Song, and Yoon}]{park2017building}
Jaehong Park, Jongyoon Song, and Sungroh Yoon. 2017.
\newblock \href {https://arxiv.org/abs/1704.00253} {Building a neural machine
  translation system using only synthetic parallel data}.
\newblock \emph{arXiv preprint arXiv:1704.00253}.

\bibitem[{Pfeiffer et~al.(2020)Pfeiffer, R{\"u}ckl{\'e}, Poth, Kamath,
  Vuli{\'c}, Ruder, Cho, and Gurevych}]{pfeiffer-etal-2020-adapterhub}
Jonas Pfeiffer, Andreas R{\"u}ckl{\'e}, Clifton Poth, Aishwarya Kamath, Ivan
  Vuli{\'c}, Sebastian Ruder, Kyunghyun Cho, and Iryna Gurevych. 2020.
\newblock \href {https://doi.org/10.18653/v1/2020.emnlp-demos.7}
  {{A}dapter{H}ub: A framework for adapting transformers}.
\newblock In \emph{Proceedings of the 2020 Conference on Empirical Methods in
  Natural Language Processing: System Demonstrations}, pages 46--54, Online.
  Association for Computational Linguistics.

\bibitem[{Philip et~al.(2020)Philip, Berard, Gall{\'e}, and
  Besacier}]{philip-etal-2020-monolingual}
Jerin Philip, Alexandre Berard, Matthias Gall{\'e}, and Laurent Besacier. 2020.
\newblock \href {https://doi.org/10.18653/v1/2020.emnlp-main.361} {Monolingual
  adapters for zero-shot neural machine translation}.
\newblock In \emph{Proceedings of the 2020 Conference on Empirical Methods in
  Natural Language Processing (EMNLP)}, pages 4465--4470, Online. Association
  for Computational Linguistics.

\bibitem[{Popovi{\'c}(2015)}]{popovic-2015-chrf}
Maja Popovi{\'c}. 2015.
\newblock \href {https://doi.org/10.18653/v1/W15-3049} {chr{F}: character
  n-gram {F}-score for automatic {MT} evaluation}.
\newblock In \emph{Proceedings of the Tenth Workshop on Statistical Machine
  Translation}, pages 392--395, Lisbon, Portugal. Association for Computational
  Linguistics.

\bibitem[{Post(2018)}]{post-2018-call}
Matt Post. 2018.
\newblock \href {https://doi.org/10.18653/v1/W18-6319} {A call for clarity in
  reporting {BLEU} scores}.
\newblock In \emph{Proceedings of the Third Conference on Machine Translation:
  Research Papers}, pages 186--191, Brussels, Belgium. Association for
  Computational Linguistics.

\bibitem[{Rasley et~al.(2020)Rasley, Rajbhandari, Ruwase, and
  He}]{rasley2020deepspeed}
Jeff Rasley, Samyam Rajbhandari, Olatunji Ruwase, and Yuxiong He. 2020.
\newblock \href {https://dl.acm.org/doi/abs/10.1145/3394486.3406703}
  {Deepspeed: System optimizations enable training deep learning models with
  over 100 billion parameters}.
\newblock In \emph{Proceedings of the 26th ACM SIGKDD International Conference
  on Knowledge Discovery \& Data Mining}, pages 3505--3506.

\bibitem[{R{\"u}ckl{\'e} et~al.(2021)R{\"u}ckl{\'e}, Geigle, Glockner, Beck,
  Pfeiffer, Reimers, and Gurevych}]{ruckle-etal-2021-adapterdrop}
Andreas R{\"u}ckl{\'e}, Gregor Geigle, Max Glockner, Tilman Beck, Jonas
  Pfeiffer, Nils Reimers, and Iryna Gurevych. 2021.
\newblock \href {https://doi.org/10.18653/v1/2021.emnlp-main.626}
  {{AdapterDrop}: {O}n the efficiency of adapters in transformers}.
\newblock In \emph{Proceedings of the 2021 Conference on Empirical Methods in
  Natural Language Processing}, pages 7930--7946, Online and Punta Cana,
  Dominican Republic. Association for Computational Linguistics.

\bibitem[{Sennrich et~al.(2016)Sennrich, Haddow, and
  Birch}]{sennrich-etal-2016-improving}
Rico Sennrich, Barry Haddow, and Alexandra Birch. 2016.
\newblock \href {https://doi.org/10.18653/v1/P16-1009} {Improving neural
  machine translation models with monolingual data}.
\newblock In \emph{Proceedings of the 54th Annual Meeting of the Association
  for Computational Linguistics (Volume 1: Long Papers)}, pages 86--96, Berlin,
  Germany. Association for Computational Linguistics.

\bibitem[{Sharaf et~al.(2020)Sharaf, Hassan, and
  Daum{\'e}~III}]{sharaf-etal-2020-meta}
Amr Sharaf, Hany Hassan, and Hal Daum{\'e}~III. 2020.
\newblock \href {https://doi.org/10.18653/v1/2020.ngt-1.5} {Meta-learning for
  few-shot {NMT} adaptation}.
\newblock In \emph{Proceedings of the Fourth Workshop on Neural Generation and
  Translation}, pages 43--53, Online. Association for Computational
  Linguistics.

\bibitem[{Swietojanski and Renals(2014)}]{swietojanski2014learning}
Pawel Swietojanski and Steve Renals. 2014.
\newblock Learning hidden unit contributions for unsupervised speaker
  adaptation of neural network acoustic models.
\newblock In \emph{2014 IEEE Spoken Language Technology Workshop (SLT)}, pages
  171--176. IEEE.

\bibitem[{Tarunesh et~al.(2021)Tarunesh, Khyalia, Kumar, Ramakrishnan, and
  Jyothi}]{tarunesh-etal-2021-meta}
Ishan Tarunesh, Sushil Khyalia, Vishwajeet Kumar, Ganesh Ramakrishnan, and
  Preethi Jyothi. 2021.
\newblock \href {https://doi.org/10.18653/v1/2021.eacl-main.314} {Meta-learning
  for effective multi-task and multilingual modelling}.
\newblock In \emph{Proceedings of the 16th Conference of the European Chapter
  of the Association for Computational Linguistics: Main Volume}, pages
  3600--3612, Online. Association for Computational Linguistics.

\bibitem[{Tiedemann(2012)}]{tiedemann-2012-parallel}
J{\"o}rg Tiedemann. 2012.
\newblock \href
  {http://www.lrec-conf.org/proceedings/lrec2012/pdf/463_Paper.pdf} {Parallel
  data, tools and interfaces in {OPUS}}.
\newblock In \emph{Proceedings of the Eighth International Conference on
  Language Resources and Evaluation ({LREC}'12)}, pages 2214--2218, Istanbul,
  Turkey. European Language Resources Association (ELRA).

\bibitem[{van~der Wees et~al.(2017)van~der Wees, Bisazza, and
  Monz}]{van-der-wees-etal-2017-dynamic}
Marlies van~der Wees, Arianna Bisazza, and Christof Monz. 2017.
\newblock \href {https://doi.org/10.18653/v1/D17-1147} {Dynamic data selection
  for neural machine translation}.
\newblock In \emph{Proceedings of the 2017 Conference on Empirical Methods in
  Natural Language Processing}, pages 1400--1410, Copenhagen, Denmark.
  Association for Computational Linguistics.

\bibitem[{Vaswani et~al.(2017)Vaswani, Shazeer, Parmar, Uszkoreit, Jones,
  Gomez, Kaiser, and Polosukhin}]{attention}
Ashish Vaswani, Noam Shazeer, Niki Parmar, Jakob Uszkoreit, Llion Jones,
  Aidan~N Gomez, \L~ukasz Kaiser, and Illia Polosukhin. 2017.
\newblock \href
  {https://proceedings.neurips.cc/paper/2017/file/3f5ee243547dee91fbd053c1c4a845aa-Paper.pdf}
  {Attention is all you need}.
\newblock In \emph{Advances in Neural Information Processing Systems},
  volume~30. Curran Associates, Inc.

\bibitem[{Vilar(2018)}]{vilar-2018-learning}
David Vilar. 2018.
\newblock \href {https://doi.org/10.18653/v1/N18-2080} {Learning hidden unit
  contribution for adapting neural machine translation models}.
\newblock In \emph{Proceedings of the 2018 Conference of the North {A}merican
  Chapter of the Association for Computational Linguistics: Human Language
  Technologies, Volume 2 (Short Papers)}, pages 500--505, New Orleans,
  Louisiana. Association for Computational Linguistics.

\bibitem[{Wolf et~al.(2020)Wolf, Debut, Sanh, Chaumond, Delangue, Moi, Cistac,
  Rault, Louf, Funtowicz, Davison, Shleifer, von Platen, Ma, Jernite, Plu, Xu,
  Le~Scao, Gugger, Drame, Lhoest, and Rush}]{wolf-etal-2020-transformers}
Thomas Wolf, Lysandre Debut, Victor Sanh, Julien Chaumond, Clement Delangue,
  Anthony Moi, Pierric Cistac, Tim Rault, Remi Louf, Morgan Funtowicz, Joe
  Davison, Sam Shleifer, Patrick von Platen, Clara Ma, Yacine Jernite, Julien
  Plu, Canwen Xu, Teven Le~Scao, Sylvain Gugger, Mariama Drame, Quentin Lhoest,
  and Alexander Rush. 2020.
\newblock \href {https://doi.org/10.18653/v1/2020.emnlp-demos.6} {Transformers:
  State-of-the-art natural language processing}.
\newblock In \emph{Proceedings of the 2020 Conference on Empirical Methods in
  Natural Language Processing: System Demonstrations}, pages 38--45, Online.
  Association for Computational Linguistics.

\bibitem[{Zhan et~al.(2021)Zhan, Liu, Wong, and Chao}]{zhan2021meta}
Runzhe Zhan, Xuebo Liu, Derek~F Wong, and Lidia~S Chao. 2021.
\newblock \href {https://www.aaai.org/AAAI21Papers/AAAI-4465.ZhanR.pdf}
  {Meta-curriculum learning for domain adaptation in neural machine
  translation}.
\newblock In \emph{Proceedings of the AAAI Conference on Artificial
  Intelligence}, volume~35, pages 14310--14318.

\bibitem[{Zhang et~al.(2020)Zhang, Williams, Titov, and
  Sennrich}]{zhang-etal-2020-improving}
Biao Zhang, Philip Williams, Ivan Titov, and Rico Sennrich. 2020.
\newblock \href {https://doi.org/10.18653/v1/2020.acl-main.148} {Improving
  massively multilingual neural machine translation and zero-shot translation}.
\newblock In \emph{Proceedings of the 58th Annual Meeting of the Association
  for Computational Linguistics}, pages 1628--1639, Online. Association for
  Computational Linguistics.

\end{thebibliography}
\bibliographystyle{acl_natbib}

\appendix
\section{Appendix}

\subsection{Datasets}
\label{appendix:dataset}
All datasets used in our experiments are publicly available on OPUS. 
Despite the fact that OPUS contains corpora from various domains and languages, some recent works \cite{aharoni-goldberg-2020-unsupervised,lai-etal-2022-improving-domain} have raised concerns about using OPUS corpora as they can be noisy.
We therefore performed the following cleaning and filtering preprocess on the original OPUS corpus:
i) remove sentences that contain more than 50\% punctuation;
ii) to ensure that the training set did not contain any corpora from the validation or test sets, all corpora were de-duplicated;
iii) sentences longer than 175 tokens were removed;
iv) we used a language detection tool\footnote{\url{https://fasttext.cc/docs/en/language-identification.html}} (\emph{langid}) to filter out sentences with mixed languages.

As described in Section~\ref{exp}, during the training phase, although most of the DLPs were limited to a maximum of 5000 sentences, there was still a fraction of DLPs with a corpus of less than 5000 samples which we list it in Table~\ref{tab:data}.

\begin{table}[h]
\resizebox{1\columnwidth}{!}{
\begin{tabular}{lr||lr}
\toprule
\textbf{DLP} & \textbf{\#Num.} & \textbf{DLP} & \textbf{\#Num.} \\
\hline
EUbookshop-hu-sr & 59 & Ubuntu-hu-sr & 140 \\
EUbookshop-hu-mk & 976 & Ubuntu-hr-sr & 438 \\
EUbookshop-en-sr & 1104 & Ubuntu-hr-hu & 479 \\
EUbookshop-et-sr & 1141 & Ubuntu-et-sr & 912 \\
EUbookshop-hr-sr & 1280 & Ubuntu-en-sr & 1519 \\
EUbookshop-mk-sr & 1320 & Ubuntu-et-mk & 1545 \\
EUbookshop-hr-hu & 1328 & Ubuntu-hr-mk & 1880 \\
EUbookshop-en-mk & 1836 & Ubuntu-mk-sr & 2091 \\
EUbookshop-et-mk & 2000 & Ubuntu-hu-mk & 2118 \\
EUbookshop-hr-mk & 2003 & Ubuntu-et-hu & 2147 \\
EUbookshop-et-hr & 2861 & Ubuntu-et-hr & 2542 \\
EUbookshop-en-hr & 4668 & Ubuntu-en-mk & 2644 \\
- & -  & Ubuntu-en-et & 4998 \\
- & - & Ubuntu-en-hu & 4999 \\   
\bottomrule
\end{tabular}}
\caption{\label{tab:data}
Data statistics (number of sentences) of DLPs that contain less than 5000 sentences.}
\end{table}

\subsection{Model Configuration}
\label{appendix:model_config}
Our \md model is trained in the following way: it first samples $m$ tasks based on temperature $\tau$, then makes $k$ gradient updates for each task $\mathcal{T}_i$. Finally, it updates the parameters of $\psi$.
In our set of experiments, we use the AdamW \cite{loshchilov2018decoupled} optimizer, which is shared across all DLPs.
We fix the initial learning rate to $5e-5$ with a dropout probability $0.1$.
In our experiments, we consider values of $m$ $\in$ \{4, 8, 16\}, $k$ $\in$ \{1, 2, 3, 4, 5\}, $\alpha$ $\in$ \{0.1,0.5,1.0\} and $\tau$ $\in$ \{1, 2, 5, $\infty$\} and choose the best setting ($m$ = 8, $k$ = 3, $\beta$ = 1.0, $\tau$ = 1) based on the average BLEU scores over all DLPs.
Each \md model is trained for 3 epochs and adapts to each DLP for 1 epoch to simulate a fast adaptation scenario.

\subsection{Additional Results}
\subsubsection{chrF Evaluation}
\label{appendix:chrf_eva}
In addition to BLEU, we also use chrF \cite{popovic-2015-chrf} as an evaluation metric. Tables~\ref{tab:main_res_chrf}, ~\ref{tab:domain_transfer_chrf} and ~\ref{tab:lang_transfer_chrf} show the results. 
\md is more effective than all baseline systems in terms of chrF, which is consistent with the BLEU scores (that were presented in Tables~\ref{tab:main_res}, ~\ref{tab:domain_transfer} and ~\ref{tab:lang_transfer}).

\subsubsection{Results on all DLPs}
\label{appendix:all_dlps}
Figure~\ref{fig:bleu_all} reports the results for all DLPs, which is consistent with the results in Tables~\ref{tab:main_res} and ~\ref{tab:main_res_chrf}.

\subsection{Analysis}
To better understand our proposed method, we investigate the effect of different parameter settings on the results (as described in Section~\ref{method:task_sampling}).
We also analyse the poor results on \textit{EUbookshop} domain as described in Section~\ref{analysis:dtvl}.

\begin{table}[t]
\centering
\begin{tabular}{lccc}
\toprule
 & \textbf{UN} & \textbf{Tanzil} & \textbf{Infopankki} \\
 \hline
$\tau=1$ & \textbf{33.53} & \textbf{9.87} & 18.43 \\
$\tau=2$ & 33.52 & 9.81 & \textbf{18.46} \\
$\tau=5$ & 33.33 & 9.77 & 18.19 \\
$\tau=\infty$ & 33.44 & 9.80 & 18.44 \\
\bottomrule
\end{tabular}
\caption{\label{tab:tempt}
Different temperature settings.
}
\end{table}

\begin{table}[t]
\centering
\begin{tabular}{lc}
\toprule
\multicolumn{1}{c}{\textbf{shots}} & \textbf{avg BLEU} \\
\hline
2-shots & 23.80 \\
4-shots & 23.88 \\
8-shots & \textbf{23.89} \\
16-shots & 23.85 \\
32-shots & 23.88 \\
\bottomrule
\end{tabular}
\caption{\label{tab:shots}
Different amounts of shots.
}
\end{table}

\subsubsection{Effect of temperature sampling}
Although the \textit{meta-training} data of all DLPs is limited to a maximum of 5000 sentences, there are still some DLPs with less than 5000 sentences, so we use temperature sampling for each setting for $\tau = $ 1, 2, 5 and $\infty$.
We first sample the task-based temperature and show the results in Table~\ref{tab:tempt}.
We notice that the performance of the various temperature settings is very similar.
These results meet our expectation as the data we used was limited to a maximum of 5000 sentences in most DLPs, with the exception of some DLPs in the \textit{EUbookshop} and \textit{Ubuntu} domains (see  Appendix~\ref{appendix:dataset}), which means data is sampled uniformly in different temperature settings.

\subsubsection{Effect of different shots}
We also test the performance on different numbers of shots ($n=2,4,6,8,16$) and show the results in Table~\ref{tab:shots}.
Interestingly, we observe that \md is not sensitive to different numbers of shots, unlike other NLP \cite{chen-etal-2022-meta} and Computer Vision tasks  \cite{finn2017model} which use the meta-learning approach.
We argue that this is because the meta-adapter is randomly initialized at each batch, resulting in a gap between training and inference.
Narrowing this gap is an important future research direction.

\subsection{Analysis on EUbookshop domain}
\label{appendix:res_EU}
\begin{figure}[t]
\centering
\subfigure[m2m + FT]{
\label{Fig.sub.1}
\includegraphics[width=0.95\columnwidth]{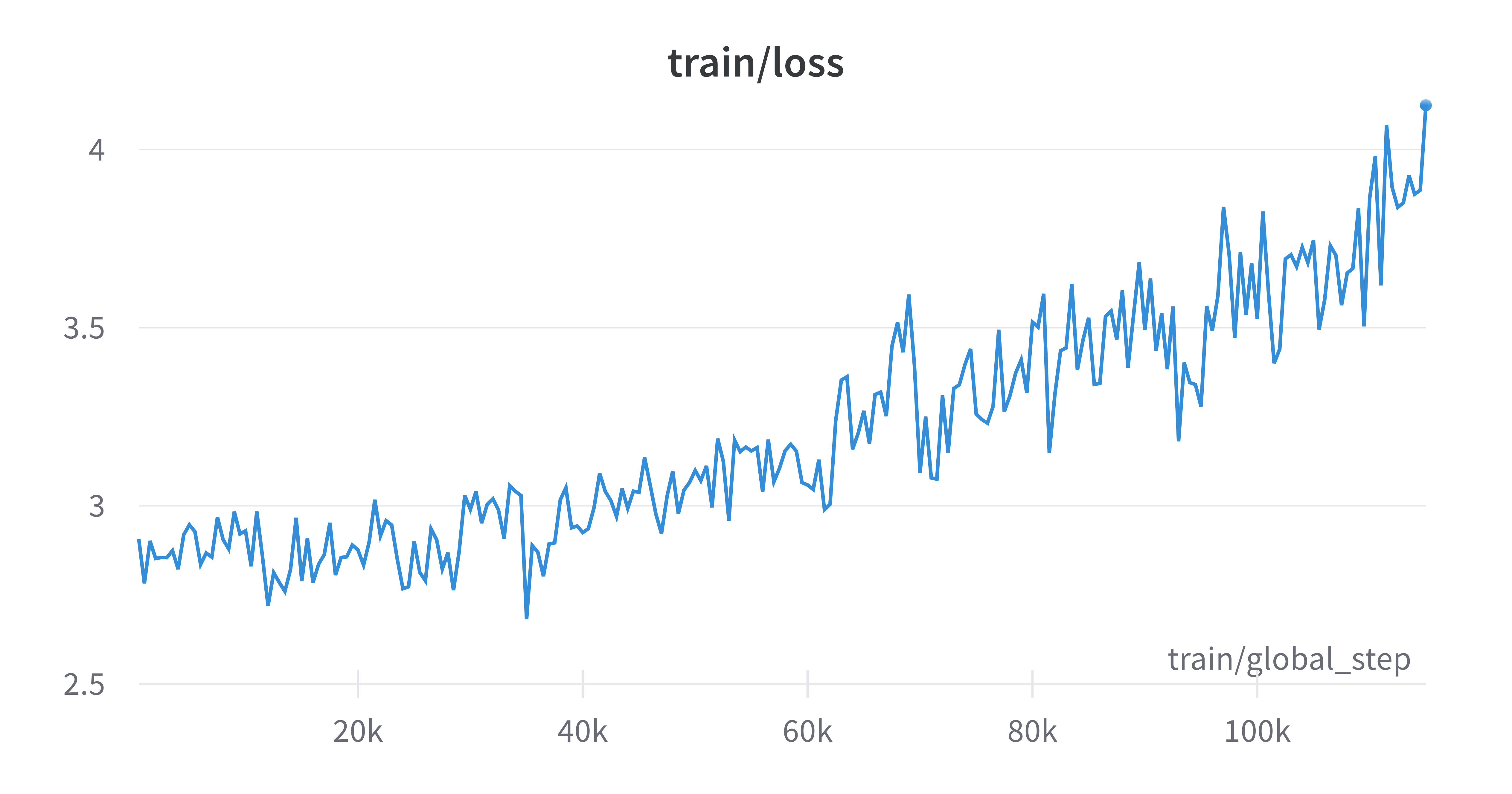}}
\subfigure[\md]{
\label{Fig.sub.2}
\includegraphics[width=0.95\columnwidth]{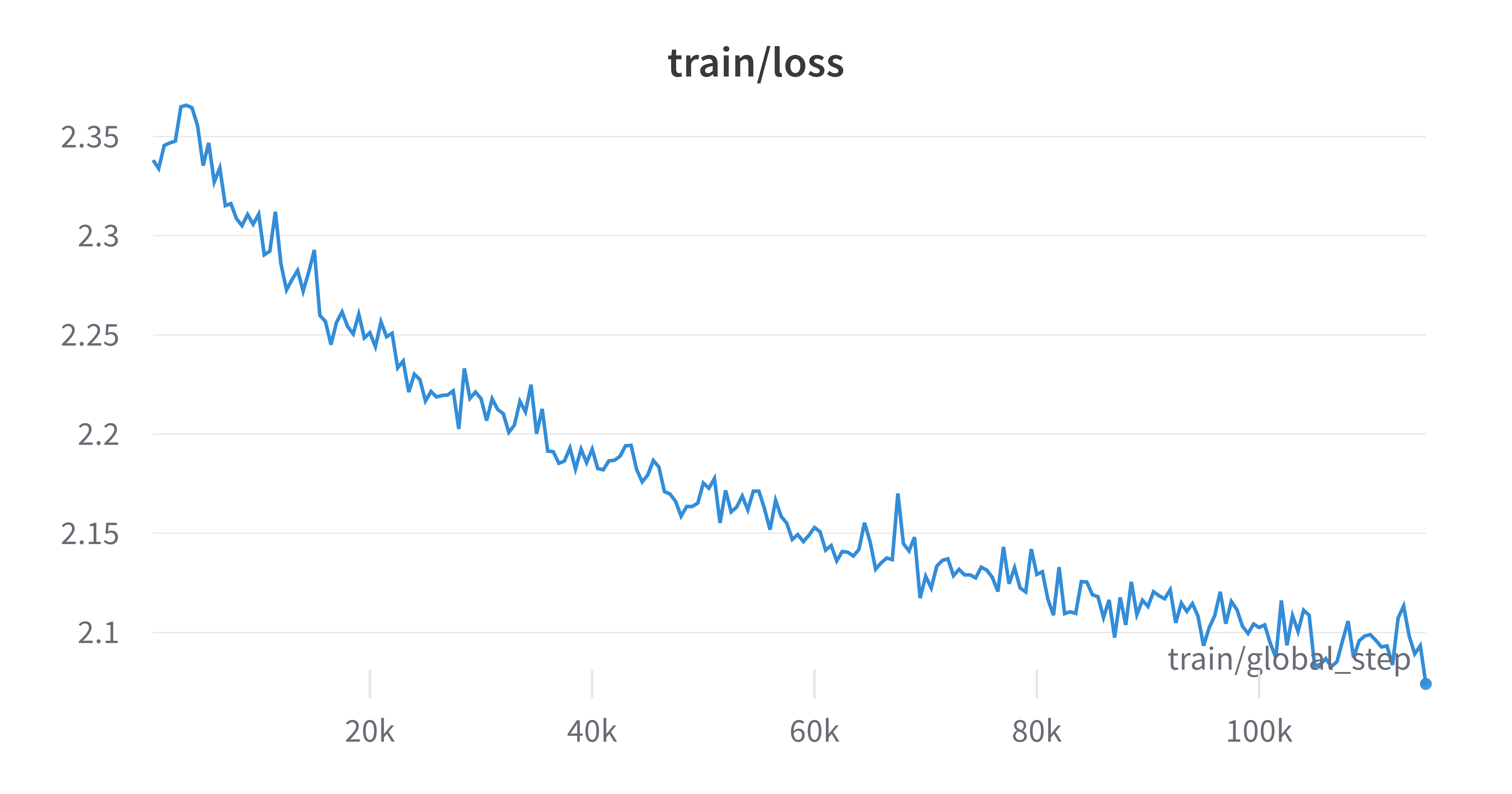}}
\caption{Training loss of \textit{m2m + FT} and \md in EUbookshop domain.}
\label{Fig.loss_EU}
\end{figure}

As described in Section~\ref{analysis:dtvl}, we observed that all baseline systems overfit when trained on data from the \textit{EUbookshop} domain.
For example, in the case of the \textit{m2m + FT} baseline, the training loss converges and stops improving at a very early stage. After that, the model overfits the validation set (Figure~\ref{Fig.loss_EU}).
On the contrary, the training loss of the \md does not show signs of overfitting. This is probably due to the much smaller number of parameters that our proposed model trains.

\begin{table*}[b]
\resizebox{\textwidth}{!}{
\begin{tabular}{lccc||ccc||ccc}
\hline
 & \multicolumn{3}{c}{\textbf{DLP (\textit{meta-adaptation} domain)}} & \multicolumn{6}{c}{\textbf{specific DLP}} \\ \hline
 & UN & Tanzil & Infopankki & UN-ar-en & Tanzil-ar-en & Infopankki-ar-en & UN-ar-ru & Tanzil-ar-ru & Infopankki-ar-ru \\ \hline
\textbf{m2m} & 0.480 & 0.227 & 0.377 & 0.602 & 0.280 & 0.479 & 0.484 & 0.191 & 0.450 \\
\textbf{m2m + FT} & 0.473 & 0.203 & 0.348 & 0.592 & 0.249 & 0.466 & 0.473 & 0.154 & 0.401 \\
\textbf{m2m + tag} & 0.473 & 0.203 & 0.344 & 0.590 & 0.255 & 0.448 & 0.474 & 0.152 & 0.400 \\
\textbf{agnostic-adapter} & 0.475 & 0.228 & 0.370 & 0.615 & 0.242 & 0.488 & 0.486 & 0.217 & 0.431 \\
\textbf{stack-adapter} & 0.472 & 0.207 & 0.368 & 0.593 & 0.243 & 0.476 & 0.473 & 0.151 & 0.405 \\
\textbf{meta-learning} & 0.487 & 0.203 & 0.349 & 0.612 & 0.278 & 0.454 & 0.483 & 0.165 & 0.428 \\ \hline
\textbf{\boldmath $m^{4}Adapter$} & \textbf{0.525} & \textbf{0.230} & \textbf{0.384} & \textbf{0.649} & \textbf{0.299} & \textbf{0.491} & \textbf{0.536} & \textbf{0.228} & \textbf{0.521} \\
\hline
\end{tabular}}
\caption{\label{tab:main_res_chrf}
Main results on the \textit{meta-adaptation stage}: average chrF scores on all DLPs with different adaptation domain (left) and chrF scores on some examples of specific DLP (right).}
\end{table*}

\begin{table*}[b]
\resizebox{\textwidth}{!}{
\begin{tabular}{lccccccc||ccccccc}
\hline
 & \multicolumn{7}{c}{\textbf{\textit{meta-adaptation} domain}} & \multicolumn{7}{c}{specific DLP (hr-sr)} \\ \hline
 & \multicolumn{1}{l}{EUbookshop} & \multicolumn{1}{l}{KDE} & \multicolumn{1}{l}{OpenSubtitles} & \multicolumn{1}{l}{QED} & \multicolumn{1}{l}{TED} & \multicolumn{1}{l}{Ubuntu} & \multicolumn{1}{l}{Bible} & EUbookshop & KDE & OpenSubtitles & QED & TED & Ubuntu & Bible \\ \hline
\textbf{m2m} & 0.446 & 0.417 & 0.339 & 0.420 & 0.408 & 0.476 & 0.129 & 0.361 & 0.432 & 0.284 & 0.204 & 0.146 & 0.495 & 0.025  \\
\textbf{m2m + FT} & 0.378 & 0.444 & 0.358 & 0.444 & 0.445 & 0.567 & 0.144 & 0.353 & 0.473 & 0.677 & 0.423 & 0.429 & 0.563 & 0.138 \\
\textbf{m2m + tag} & 0.388 & 0.445 & 0.359 & 0.414 & 0.428 & 0.520 & 0.135 & 0.359 & 0.502 & 0.671 & 0.360 & 0.428 & 0.581 & 0.118  \\
\textbf{agnostic-adapter} & 0.419 & \textbf{0.460} & 0.385 & 0.456 & \textbf{0.461} & \textbf{0.568} & 0.144  & 0.279 & \textbf{0.507} & 0.613 & 0.388 & 0.415 & 0.554 & 0.127  \\
\textbf{stack-adapter} & 0.382 & 0.436 & 0.390 & 0.438 & 0.441 & 0.546 & 0.134  & 0.358 & 0.427 & 0.562 & 0.381 & 0.427 & 0.526 & 0.124  \\
\textbf{meta-learning} & 0.387 & 0.440 & 0.360 & 0.412 & 0.422 & 0.509 & 0.142 & 0.237 & 0.502 & 0.676 & 0.353 & 0.404 & 0.546 & 0.139  \\ \hline
\textbf{\boldmath $m^{4}Adapter$} & \textbf{0.497} & 0.452 & \textbf{0.386} & \textbf{0.456} & 0.457 & 0.565 & \textbf{0.148} & \textbf{0.369} & 0.504 & \textbf{0.679} & \textbf{0.427} & \textbf{0.431} & \textbf{0.578} & \textbf{0.143} \\
\hline
\end{tabular}}
\caption{Domain transfer via languages: average chrF scores on all DLPs in each \textit{meta-adaptation} domain (left) and chrF scores on random select one specific DLP in \textit{hr-sr} (right).}
\label{tab:domain_transfer_chrf}
\end{table*}

\begin{table*}[b]
\resizebox{\textwidth}{!}{
\begin{tabular}{lcccc||cccccc}
\hline
 & \multicolumn{4}{c}{\textbf{\textit{meta-adaptation} language pair}} & \multicolumn{6}{c}{\textbf{specific DLP(de-en)}} \\ \hline
 & de-en & en-fr & fi-uk & is-lt & EUbookshop & KDE & OpenSubtitles & QED & TED & Ubuntu \\ \hline
\textbf{m2m} & 0.116 & 0.130 & 0.327 & 0.320 & 0.171 & 0.104 & 0.093 & 0.107 & 0.132 & 0.095 \\
\textbf{m2m + FT} & 0.112 & 0.094 & 0.253 & 0.243 & 0.164 & 0.091 & 0.089 & 0.105 & 0.134 & 0.090 \\
\textbf{m2m + tag} & 0.094 & 0.096 & 0.258 & 0.261 & 0.140 & 0.067 & 0.082 & 0.088 & 0.116 & 0.077 \\
\textbf{agnostic-adapter} & 0.116 & 0.127 & \textbf{0.343} & 0.331 & 0.168 & 0.102 & 0.093 & 0.108 & 0.134 & 0.092 \\
\textbf{stack-adapter} & 0.113 & 0.096 & 0.256 & 0.258 & 0.164 & 0.087 & 0.088 & 0.105 & 0.130 & 0.075 \\
\textbf{meta-learning} & 0.115 & 0.125 & 0.317 & 0.309 & 0.170 & 0.101 & 0.092 & 0.108 & 0.133 & 0.092 \\ \hline
\boldmath \md & \textbf{0.117} & \textbf{0.131} & 0.342 & \textbf{0.333} & \textbf{0.174} & \textbf{0.107} & \textbf{0.095} & \textbf{0.108} & \textbf{0.134} & \textbf{0.097} \\
\hline
\end{tabular}}
\caption{Language transfer via domains: average chrF scores on all DLPs in each \textit{meta-adaptation} language pair (left) and chrF scores on one specific DLP in \textit{de-en} (right).}
\label{tab:lang_transfer_chrf}
\end{table*}

\begin{figure*}
	\centering
	\subfigure[m2m]{
		\begin{minipage}[b]{1\textwidth}
			\includegraphics[width=1\textwidth]{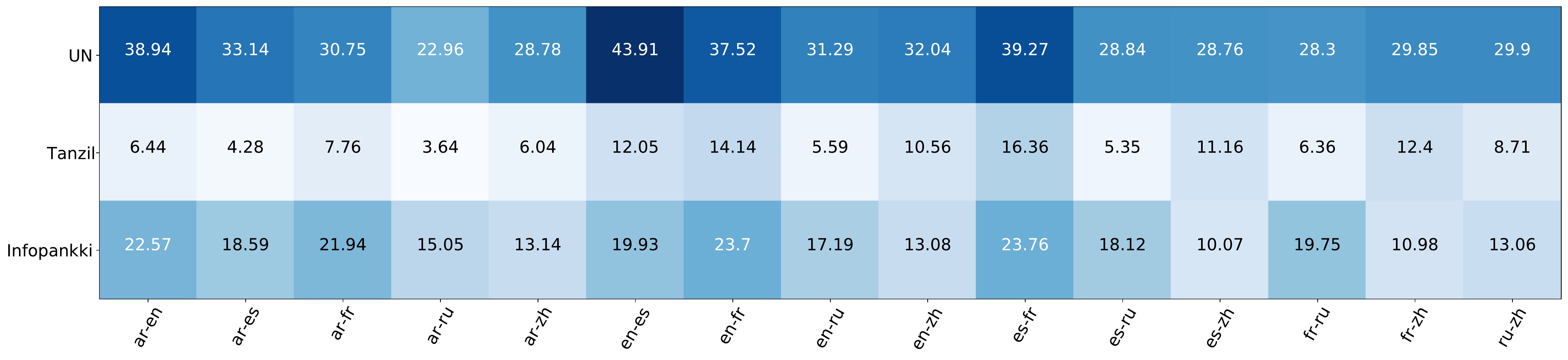} 
		\end{minipage}
	}
    	\subfigure[m2m + FT]{
    		\begin{minipage}[b]{1\textwidth}
      	 	\includegraphics[width=1\textwidth]{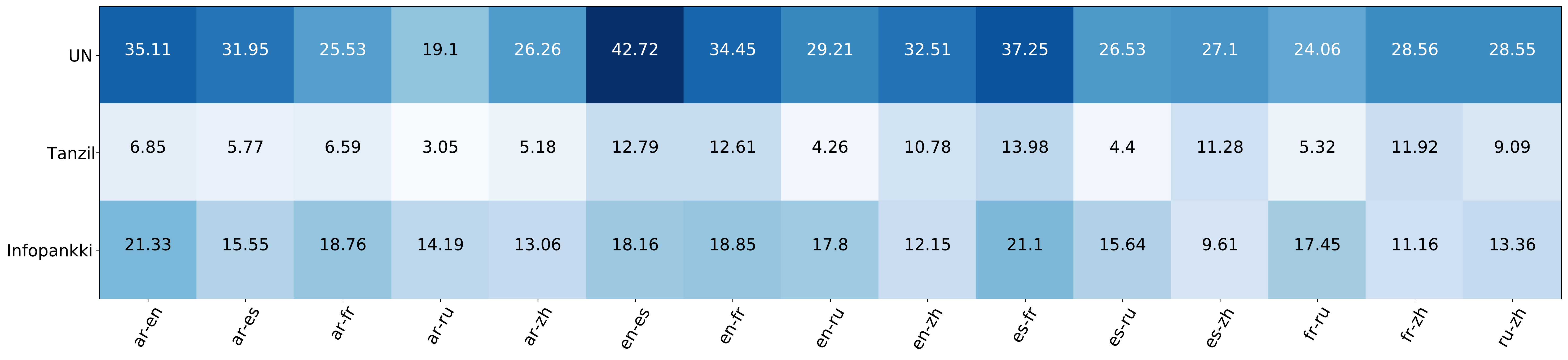}
    		\end{minipage}
    	}
	\subfigure[m2m + tag]{
		\begin{minipage}[b]{1\textwidth}
			\includegraphics[width=1\textwidth]{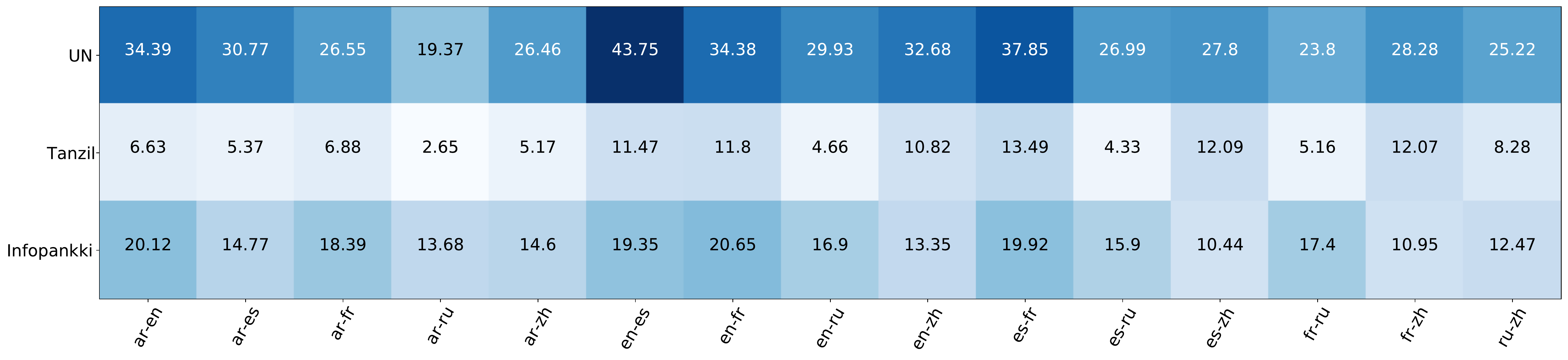} 
		\end{minipage}
	}
    	\subfigure[agnostic-adapter]{
    		\begin{minipage}[b]{1\textwidth}
		 	\includegraphics[width=1\textwidth]{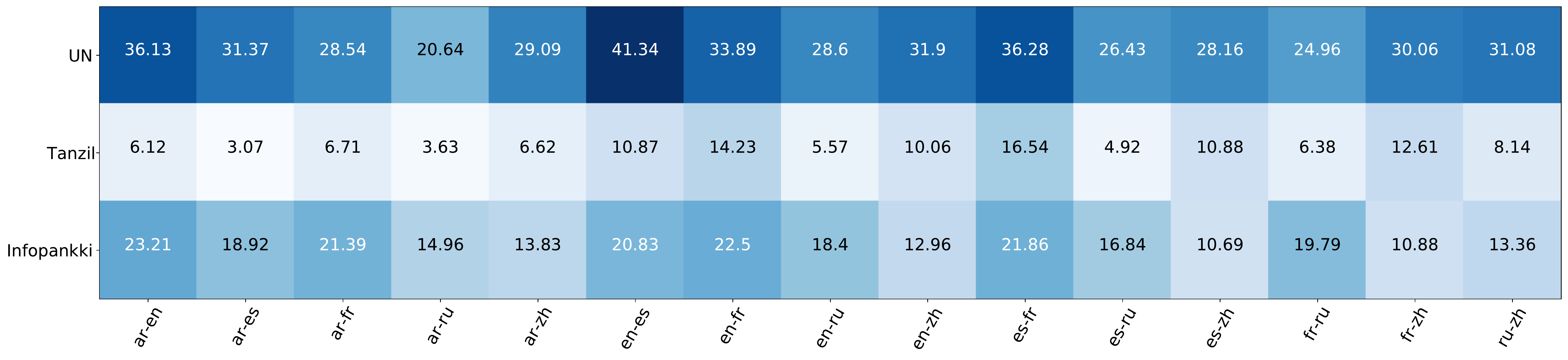}
    		\end{minipage}
    	}
    \subfigure[stack-adapter]{
		\begin{minipage}[b]{1\textwidth}
			\includegraphics[width=1\textwidth]{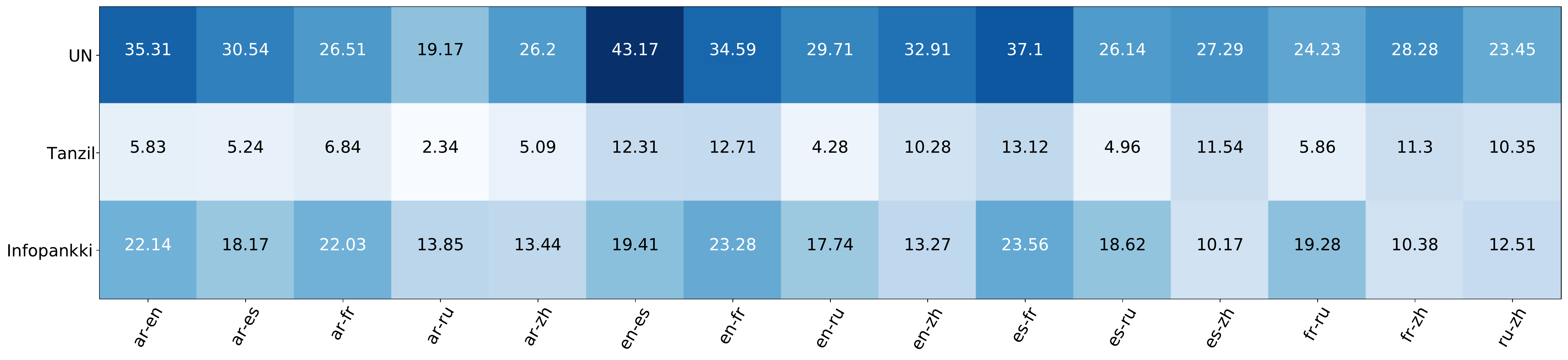} 
		\end{minipage}
	}
	\end{figure*}
\begin{figure*}[t]
    \centering
    \ContinuedFloat
	\subfigure[Meta-Learning]{
		\begin{minipage}[b]{1\textwidth}
	 	\includegraphics[width=1\textwidth]{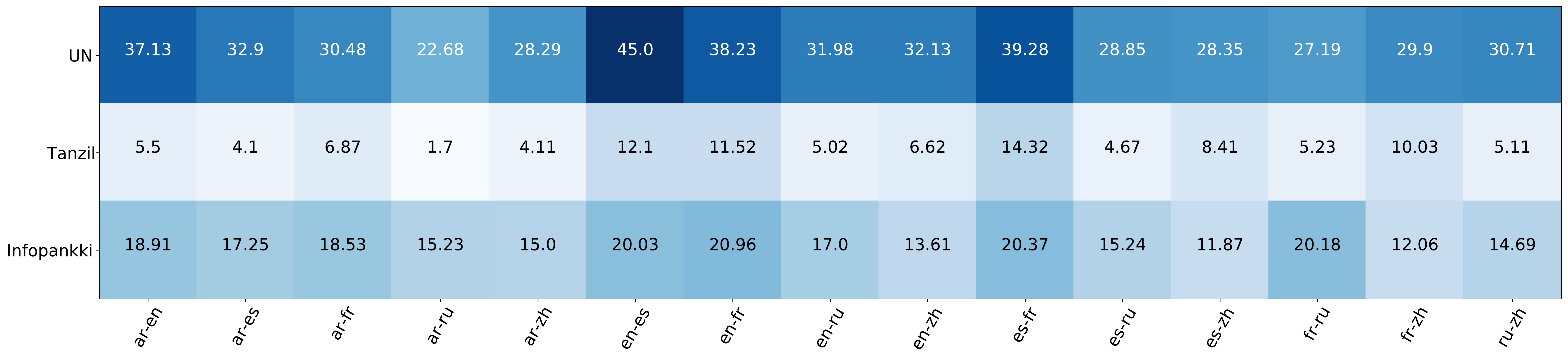}
		\end{minipage}
	}
 \subfigure[\md]{
	\begin{minipage}[b]{1\textwidth}
		\includegraphics[width=1\textwidth]{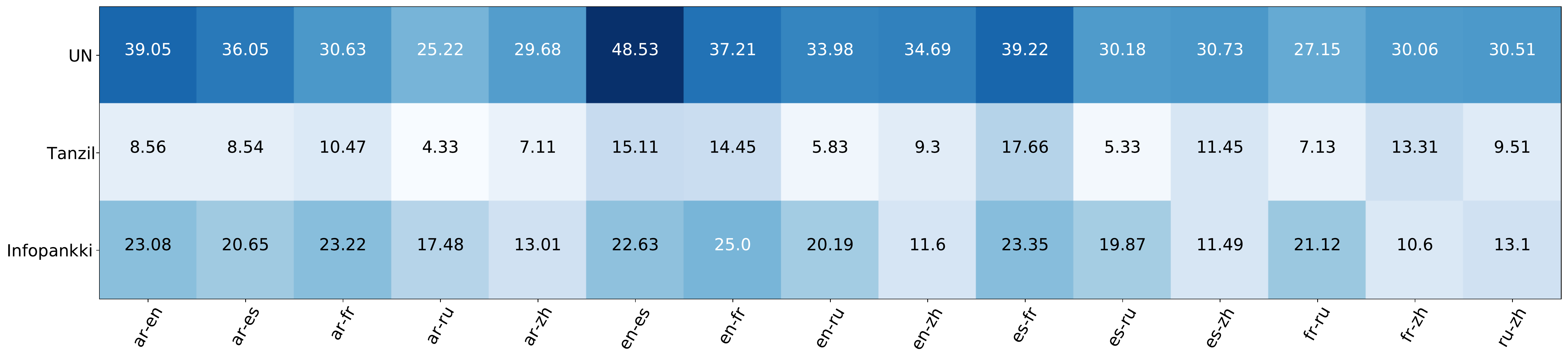} 
	\end{minipage}
}
\caption{Main result: BLEU scores in all DLPs}
\label{fig:bleu_all}
\end{figure*}

\end{document}